\documentclass[10pt,twocolumn,twoside]{IEEEtran}
\usepackage[utf8]{inputenc}
\usepackage{enumitem}    % more compact lists
\usepackage{eurosym}
\usepackage{color,soul} % only for revisions of the template 
\usepackage{url}
\usepackage{graphicx}
\usepackage{amsmath,amsfonts,amssymb,amscd,amsthm,xspace}
\usepackage{import}
\usepackage{xcolor}
\usepackage{array}
\usepackage{ragged2e}
\usepackage{subfig}
\usepackage{mathtools}
\usepackage{math}
\usepackage{multirow}
\usepackage{enumitem}
\usepackage{cite}
\usepackage{multirow}
\usepackage{booktabs}
\usepackage[lined,boxed,commentsnumbered,ruled,norelsize]{algorithm2e}
\setlist{nolistsep}
\usepackage
  [bookmarks=true,bookmarksopen=true,colorlinks=true,linkcolor=blue,
   citecolor=blue,urlcolor=blue]
  {hyperref}

\title{Face Frontalization Based on Robustly Fitting a Deformable Shape Model to 3D Landmarks
}
\author{Zhiqi Kang, Mostafa Sadeghi and Radu Horaud \thanks{Z. Kang, M. Sadeghi and R. Horaud are with Inria Grenoble and with Universit\'e Grenoble Alpes.}}

\date{}
\begin{document}
\maketitle

\begin{abstract}
Face frontalization consists of synthesizing a frontally-viewed face from an arbitrarily-viewed one. The main contribution of this paper is a robust face alignment method that enables pixel-to-pixel warping. The method simultaneously estimates the \textit{rigid transformation} (scale, rotation, and translation) and the \textit{non-rigid deformation} between two 3D point sets: a set of 3D landmarks extracted from an arbitrary-viewed face, and a set of 3D landmarks parameterized by a frontally-viewed deformable face model. An important merit of the proposed method is its ability to deal both with noise (small perturbations) and with outliers (large errors). We propose to model inliers and outliers with the generalized Student's t-probability distribution function -- a heavy-tailed distribution that is immune to non-Gaussian errors in the data. We describe in detail the associated expectation-maximization (EM) algorithm that alternates between the estimation of (i)~the rigid parameters, (ii)~the deformation parameters, and (iii)~ the t-distribution parameters. 
We also propose to use the \textit{zero-mean normalized cross-correlation}, between a frontalized face and the corresponding ground-truth frontally-viewed face,  to evaluate the performance of frontalization. To this end, we use a dataset that contains pairs of profile-viewed and frontally-viewed faces. This evaluation, based on direct image-to-image comparison, stands in contrast with indirect evaluation, based on analyzing the effect of frontalization on face recognition.\footnote{Supplemental material for this paper is accessible at \url{https://team.inria.fr/perception/research/rff/}.}

\end{abstract}

\section{Introduction}
\label{sec:introduction}
% Introduction

The problem of face frontalization is the problem of synthesizing a frontal view of a face from an arbitrary view. Recent research has shown that face frontalization consistently boosts the performance of face analysis. In particular, it has been recently demonstrated that face recognition from frontal views yields better performance than face recognition from unconstrained views \cite{yim2015rotating,banerjee2018frontalize,zhao2018towards}. This observation is equally valid for other tasks, such as the analysis of facial expressions \cite{pei2020monocular} or of lip reading \cite{fernandez2018survey,adeel2019lip}. %These analyses often imply video processing rather than single-image processing. %Obviously, inherent rigid head motions strongly perturb the temporal understanding of non-rigid facial movements. 

It is well established that lip movements, as well as tongue and jaw movements, are controlled by speech production and that they are correlated with word pronunciation \cite{schultz2017biosignal}. Consequently, the capacity to properly analyse facial movements plays an important role in visual and audio-visual speech separation, speech enhancement and speech recognition, e.g. \cite{dupont2000audio,rivet2006mixing,wu2016novel,sadeghi2020audio,tao2020end}. This is particularly useful when audio signals are corrupted by ambient noise and by acoustic perturbations. Nevertheless, facial-movement analysis is perturbed by inherent rigid head movements. 
It is therefore important to separate non-rigid facial movements from rigid head movements, and face frontalization may well be viewed as such a rigid/non-rigid separation process. 

\begin{figure*}[t!]
\centering
 \includegraphics[width=1.01\linewidth]{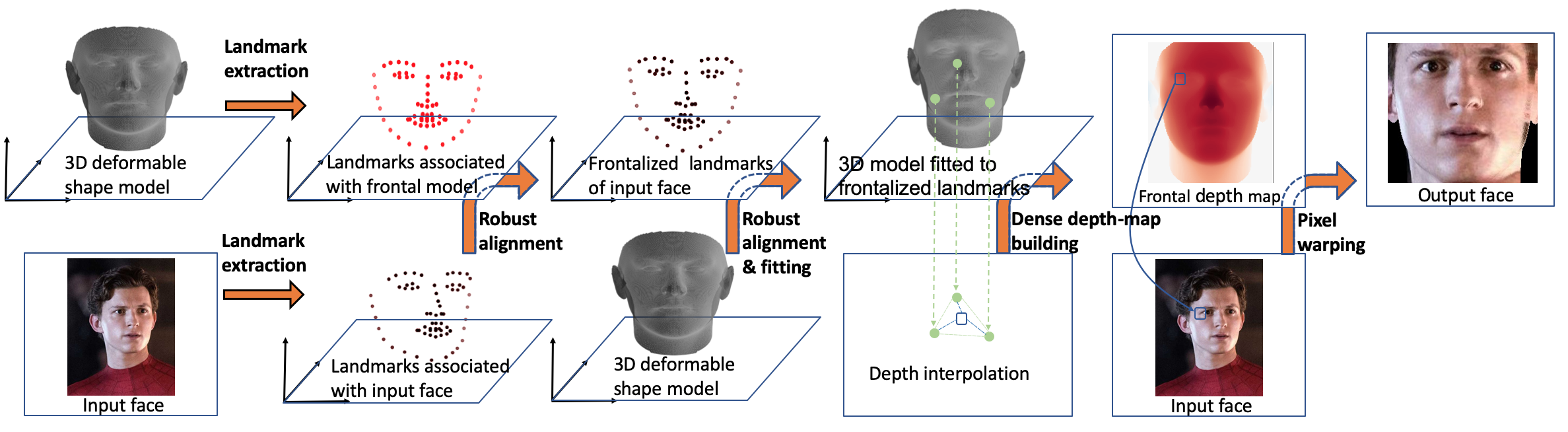}
\caption{\label{fig:pipeline} Overview of the proposed robust face frontalization method. 3D landmarks are extracted from both an arbitrarly-viewed input face and from a frontally-viewed deformable shape model. These landmarks are robustly aligned, thus enabling to compute the pose of the input face and to frontalize the input landmarks. The deformable shape model is next fitted to these landmarks, thus obtaining a frontalized shape model, followed by the computation of a frontal dense depth map; the latter is obtained by interpolation of the 3D vertices of the triangulated mesh that describes the shape model.  Finally, the input-face pixels are warped onto the output-image pixels.}
\end{figure*}

In this paper we address face frontalization as the problem of simultaneously estimating the 3D pose and deformation parameters of an arbitrarily-viewed face, e.g. Figure~\ref{fig:pipeline}. The main contribution is a robust point-set alignment method that enables face frontalization via pixel-to-pixel warping. 
In details, we propose a method that simultaneously estimates the \textit{rigid transformation} (scale, rotation, and translation) and the \textit{non-rigid deformation} between two 3D point sets, namely (i)~a set of 3D landmarks extracted from an arbitrary-viewed face and (ii)~a set of frontally-viewed 3D landmarks that are parameterized by a deformable face model. An important merit of the proposed alignment estimator is its ability to deal both with noise (small perturbations) and with outliers (large errors), i.e. robust estimation. We propose to model both inliers and outliers with the generalized Student's t-probability distribution function (pdf) -- a heavy-tailed distribution that is immune to non-Gaussian errors in the data \cite{sun2010robust,forbes2014new}. We describe in detail the associated expectation-maximization (EM) algorithm that alternates between the estimation of (i)~the rigid parameters, (ii)~the deformation parameters, and (iii)~ the pdf parameters. Interestingly, the proposed method can be indifferently applied to align two rigid point sets or to \textit{fit and align} a deformable point set to a rigid one.

We also propose to use the \textit{zero-mean normalized cross-correlation} (ZNCC) coefficient between a frontalized face and the corresponding ground-truth frontally-viewed face,  in order to empirically evaluate the performance of face frontalization. To this end, we use a dataset that contains pairs of profile-viewed and frontally-viewed faces. This direct evaluation, based on image-to-image comparison between the prediction and the ground-truth, stands in contrast with indirect evaluation, based on analyzing the effect of frontalization on face recognition, e.g. \cite{yim2015rotating,banerjee2018frontalize,zhao2018towards}.

The remainder of this article is organized as follows. In Section~\ref{sec:related-work} we review face frontalization methods that were proposed in the recent past. Section~\ref{sec:frontalization-landmarks} outlines the proposed algorithm and describes in detail the proposed 3D point-set alignment method. Section~\ref{sec:implementation} analyzes some technical aspects of the method and provides implementation details. Section~\ref{sec:experiments} describes the proposed evaluation metrics and shows results obtained with our method and with two state-of-the-art methods. Section~\ref{sec:conclusions} draws some conclusions. Finally, Appendix~\ref{app:robust-alignment}, Appendix~\ref{app:statistical-shape} and Appendix~\ref{app:robust-shape-fit} provide mathematical details of the proposed algorithms.

\section{Related Work}
\label{sec:related-work}
%related-work

%The following topics should be discussed:
%
%\begin{itemize}
%\item face frontalization
%\item deformable shape models and how they are fit to the data
%\item robust point-set registration
%\item benchmarking face frontalization
%\end{itemize}

As already mentioned, face frontalization consists of synthesizing a frontally-viewed face from an arbitrarily-viewed face. 
A popular approach has been to train deep neural networks (DNNs) to learn a mapping from an arbitrary view to a frontal view, using a massive collection of input/target pairs of faces.  It was shown that architectures based on generative adversarial networks (GANs), e.g.  \cite{huang2017beyond,zhao2018towards,tran2017disentangled,rong2020feature} outperform architectures based on convolutional neural networks (CNNs) \cite{yim2015rotating} for face recognition. The main drawback of these DNN-based methods is that they are designed to predict frontal faces that are as neutral as possible, in order to improve the performance of face recognition. 
Hence, there is no guarantee that the predicted output preserves non-rigid facial deformations. These methods are therefore hardly usable for facial expression recognition or for lip reading.

%The main motivation is to improve the performance of face recognition. 
Another approach has been to estimate the 3D pose of an input face with respect to a frontal 3D face model, and then to use the pose parameters to warp the facial pixels from the input image onto a frontal view. Recently proposed face frontalization methods capitalize on 3D pose estimation, e.g. \cite{hassner2015effective}, \cite{banerjee2018frontalize}. The problem of 3D pose estimation from point-to-point correspondences has been thoroughly addressed in the past and many solutions have been proposed in the presence of various camera models, e.g. \cite{horaud1997object}.  In \cite{hassner2015effective} it was proposed to use a 3D generic model of a face from which a frontal image is synthesized. Next, 48 2D facial landmarks are extracted from the input face and from the generic frontal face, which provides 2D-to-3D correspondences between the input face and the generic 3D model. This amounts to estimate the projection (or camera) matrices, from the 3D face model onto the input image and onto the frontal image, respectively, and to subsequently warp pixels from one image to another. A similar method was proposed in \cite{banerjee2018frontalize} where 68 2D landmarks are extracted from both the input face and from a frontal face associated with a generic 3D model. The 3D pose parameters are obtained from 2D-to-3D landmark correspondences between the input face and the 3D model.  

Nevertheless, the pose-based methods just cited suffer from a number of limitations. First, the 3D models that they use do not take facial deformations into account. Second, the estimation of 3D pose with an unknown projective camera model is an ill-posed problem in the case of quasi-planar shape models such as faces. Third, the pose estimator itself is not robust to the presence of large errors in landmark detection and localization, and to large non-rigid facial deformations. 

The proposed method overcomes these limitations on the following grounds. First, we use a deformable 3D face model that parameterizes faces with large variabilities in appearance and in expression. Deformable shape models were initially introduced in \cite{cootes1995active}, and subsequently used in \cite{blanz1999morphable,paysan20093d,cao2013facewarehouse} to model faces and facial deformations. Second, we use 3D facial landmarks rather than 2D landmarks, e.g. \cite{bulat2017far,feng2018joint,zhu2019face}. Consequently, there is no need to estimate the parameters of the camera model, at the price of approximating the latter with a scaled orthographic (or weak perspective) model. However, the latter is a realistic approximation of the true camera model, since human faces are quasi planar surfaces and, hence, their depth range is small relative to the camera-to-face distance \cite{horaud1997object}.  Third, we propose a robust method for aligning a 3D set of landmarks with a 3D deformable face model. The associated EM solver alternates between the estimation of the rigid-transformation parameters, the model-deformation parameters, and the robust statistical-model parameters. This paper builds on a recent study showing that the use of 3D facial landmarks in combination with robust pose estimation yields excellent performance \cite{sadeghi2020unsupervised}.

As already mentioned, face frontalization has been essentially used as a pre-processing step for face recognition and the evaluation metrics proposed so far are generally based on the recognition rate \cite{yim2015rotating,wang2016facial,banerjee2018frontalize,zhao2018towards}. This kind of evaluation lacks direct significance for standalone face frontalization methods. Moreover, existing evaluation pipelines include facial-expression normalization \cite{zhu2015high}, which biases the results and does not allow to evaluate the extent to which face frontalization preserves non-rigid facial deformations.

%\subsection{Benchmarking face frontalization}
%To the best of our knowledge, there is not yet a benchmark for quantitatively evaluating the quality of frontalization, especially in the context of videos. The exiting evaluation is either in a quantitative manner or in an embedding manner, or a mixture of those two \cite{wang2016facial}. The quantitative evaluation is straightforward but easily biased, since we cannot display all results in a publication. Thus, whether the selected examples are representative to the actual performance remains problematic. An embedding approach is to use face frontalization as a preprocessing of other tasks, such as face recognition \cite{yim2015rotating, banerjee2018frontalize, zhao2018towards}, face verification \cite{hassner2015effective}, etc. This evaluation is usually quantitative, but lack of direct significance regarding the frontalization quality. In many circumstances, the existing frontalization methods includes normalization of expression, illumination. This makes it more questionable to use the substantial result (i.e. recognition rate) as a measure of the quality of frontalization. In this work, our goal is to benchmark the face frontalization by using a well-constructed video data set and a novel metric for evaluation in face frontalization. 

\section{Robust Face Frontalization}
\label{sec:frontalization-landmarks}
\subsection{Outline of the Proposed Method}
\label{sec:outline}
% outline

The proposed robust face frontalization (RFF) method is summarized in Algorithm~\ref{algo:face-frontalization} and Figure~\ref{fig:pipeline}. The core idea of the method is to estimate the 3D pose (scale, rotation, and translation) of an input face based on robust alignment of two 3D point sets, namely a set of 3D landmarks extracted from the input face, and a set of 3D landmarks associated with the frontal pose of a 3D shape model. This is done in the framework of maximum likelihood estimation and, in practice, an expectation-maximization algorithm is being used, i.e. \eqref{eq:expectation-student} and Appendix~\ref{app:robust-alignment}. This allows to frontalize the input landmarks, i.e. \eqref{eq:landmark-frontalization} and, then, to fit a deformable shape model to the frontalized landmarks, which yields a frontalized shape model of the input face, i.e. \eqref{eq:frontal-vertices}. Next, vertex interpolation is used to compute a dense depth map associated with the frontalized shape model, i.e. \eqref{eq:barycenter} and \eqref{eq:weighted-interpolation}. Eventually, a frontalized face is computed by warping the input face onto a frontal image, i.e. \eqref{eq:pixel-pred} and \eqref{eq:face-warping}.

\begin{algorithm}[t!]
 \caption{\label{algo:face-frontalization} Robust face frontalization (RFF).}

 \KwIn{Face image and deformable 3D shape model }
 \textbf{Initialization:} Extract landmarks from the input face. Extract frontal landmarks from the shape model \;
 \textbf{Pose estimation:} Compute the scale, rotation and translation parameters between the input landmarks and the model landmarks \;
 \textbf{Landmark frontalization:} Apply the pose parameters to the input landmarks \;
 \textbf{Model fitting:} Align the shape model (scale, rotation, translation and deformation parameters) with the frontalized landmarks and frontalize the shape model~\;
 \textbf{Frontal dense depth map:}  Interpolate the frontalized shape model \;
 \textbf{Face warping:} Compute pixel-to-pixel correspondences between input and output face images \;
 \KwOut{Frontalized face image}
\end{algorithm}
\subsection{Frontalization via Alignment of 3D Point Sets}

Let $I_p$ be an \textit{observed} image of a face in an unknown pose.\footnote{The term \textit{face pose} refers to a rigid transformation, namely the scale factor, rotation matrix and translation vector that map a face-centered coordinate frame onto a coordinate frame that is aligned with the camera frame.} A set $\mathcal{X}$ of $J=68$ 3D landmarks is extracted from $I_p$ with image-centered coordinates $\Xvect_{1:J} = \{ \Xvect_j \}_{j=1}^J \subset \mathbb{R}^{3}$. Throughout the paper we adopt the notation $\Xvect_j=(X_{j1}, X_{j2}, X_{j3})$ to designate the three coordinates of a point in $\mathbb{R}^3$.
Let $\Zvect_{1:J} = \{ \Zvect_j \}_{j=1}^J \subset \mathbb{R}^{3}$ be the 3D coordinates of a set of landmarks, $\mathcal{Z}$, that correspond to the frontal pose of a 3D deformable face model, e.g. \cite{sadeghi2020unsupervised}, and let $I_f$ be the frontal image corresponding to this frontal view of the model. The problem of face-pose estimation consists of finding the rigid transformation that best maps $\Xvect_{1:J}$ onto $\Zvect_{1:J}$ on the premise that the landmarks are in one-to-one correspondence and up to an error (or residual) $\evect_j$:
\begin{equation}
\label{eq:rigid-residual}
\evect_j  =  \Zvect_j -  (\rho \Rmat \Xvect_j + \tvect), \quad \forall j \in \{1 \dots J\},
\end{equation}
where $\rho\in\mathbb{R}^+$, $\Rmat\in\mathbb{R}^{3\times 3}$ and $\tvect\in\mathbb{R}^3$ are the scale, rotation matrix and translation vector, respectively, associated with the unknown face pose. Because the landmark locations $\Xvect_{1:J}$ are inherently affected by detection noise and outliers, as well as by \textit{non-rigid facial deformations}, it is suitable to use a robust alignment technique. For this purpose, we assume that the residuals $\evect_{1:J}$ are samples of a random variable $\evect$ drawn from a robust probability distribution function (pdf) $P(\evect;\thetavect)$, where $\thetavect$ is the set of parameters characterizing the pdf. Then, the problem is cast into maximum likelihood estimation (MLE) or, equivalently into the minimization of the following negative log-likelihood function:
\begin{equation}
\label{eq:neg-log-L}
\mathcal{L} (\thetavect, \rho, \Rmat, \tvect | \Xvect_{1:J}, \Zvect_{1:J}) = - \frac{1}{2} \sum_{j=1}^J \log P (\evect_j; \thetavect),
\end{equation}
The generalized Student's t-distribution \cite{forbes2014new} is a robust pdf, namely:
\begin{align}
\label{eq:generalized-student}
P(\evect; \thetavect)   = \int_{0}^{\infty} \mathcal{N} ( \evect; 0, w \inverse \Sigmamat) \mathcal{G}(w; \mu, \nu) dw %\nonumber 
%&=
%=
%\frac{\Gamma(\mu + \frac{3}{2})}{| \Sigmamat |^{\frac{1}{2}} \Gamma(\mu) (2\pi \nu)^{\frac{3}{2}}}
%\left(
%1+ \frac{\| \evect \|^2_{\Sigmamat}} {2\nu}
%\right)_{,}^{-\left(\mu + \frac{3}{2}\right)}
\end{align}
where $\mathcal{N}()$ is the normal distribution, $\mathcal{G} ()$ is the gamma distribution and $\Sigmamat\in\mathbb{R}^{3\times 3}$ is a covariance matrix. $w\in\mathbb{R}$ is a precision associated with a residual. The precisions $w_{1:J}$ are modeled as random variables drawn from gamma distributions.%parameterized by $\mu_{1:J}$ and $\nu=1$.  
%The precision weights the relative importance of a landmark: the larger the better. 

Direct minimization of \eqref{eq:neg-log-L} is not practical. An expectation-maximization (EM) formulation is therefore adopted, namely minimization of the \textit{expected complete-data negative log-likelihood}:
\begin{equation}
\label{eq:expected-student}
\min_{\thetavect} \mathrm{E}_W [ - \log P( \evect_{1:J}, w_{1:J} | \evect_{1:J} ; \thetavect)],
\end{equation}
where EM alternates between the estimation of the means of the precision posteriors, $\overline{w}_{1:J}$ and the estimation of the parameters $\thetavect = \{\rho, \Rmat, \tvect, \Sigmamat, \mu\}$. As it can be seen in Appendix~\ref{app:robust-alignment}, the precisions are precious because they weight the relative importance of the landmarks within the optimization process, and hence they help the parameter estimation process to be robust against badly localized landmarks.
Once the pose parameters are estimated, the rigid transformation is applied to the landmarks $\Xvect_{1:J}$ in order to obtain a set $\mathcal{Y}$ of \textit{frontalized} landmarks whose coordinates in $I_f$ are denoted $\Yvect_{1:J}\subset \mathbb{R}^{3}$, namely:
\begin{equation}
\label{eq:landmark-frontalization}
\Yvect_j = \rho \Rmat \Xvect_j + \tvect, \quad \forall j \in \{1 \dots J\}.
\end{equation}

The next step is to fit a deformable 3D face-shape model to this set of landmarks in order to eventually obtain a \textit{frontal dense depth map} of the face. For that purpose and without loss of generality, we consider a linear deformation model, e.g. the 3DMM Basel Face Model (BFM) \cite{paysan20093d}. The latter consists of a 3D mesh with a set $\hat{\mathcal{V}}$ of $N$ vertices, whose coordinates $\hat{\Vvect}_{1:N} = \{ \hat{\Vvect}_n \}_{n=1}^N \subset \mathbb{R}^{3}$ are parameterized by a \textit{statistical linear shape-model} in the following way (please consult Appendix~\ref{app:statistical-shape}):
\begin{equation}
\hat{\Vvect}_n = \overline{\Vvect}_n + \Wmat_n \svect, \quad \forall n \in \{1 \dots N\},
\end{equation}
where $\overline{\Vvect}_{1:N}\subset \mathbb{R}^{3}$ are the vertices of the mean shape model, $\Wmat_{1:N}\subset \mathbb{R}^{3\times K}$  are reconstruction matrices, and $\svect \in \mathbb{R}^{K}$ is a low dimensional embedding of the vertex set, with $K \ll 3N$, see  Appendix~\ref{app:statistical-shape}. In order to fit this deformable 3D face-shape model to the frontalized landmarks $\Yvect_{1:J}$, we consider a subset of $J=68$ vertices with coordinates $\hat{\Vvect}_{1:J}$  that correspond, one-to-one, to the frontalized landmarks, namely $\{\hat{\Vvect}_{j} \leftrightarrow \Yvect_{j}\}_{j=1}^J$.\footnote{For the sake of simplifying the notations, we use  $\{1\dots J\}\subset\{1\dots N\}$.} For that purpose, the statistical shape model must be scaled, rotated, translated and deformed, such that the vertices $\hat{\Vvect}_{1:J}$ are optimally aligned with the landmarks $\Yvect_{1:J}$. This yields the following negative log-likelihood function:
\begin{equation}
\label{eq:neg-log-shape}
\mathcal{L} (\thetavect, \sigma, \Qmat, \dvect, \svect | \Yvect_{1:J}, \hat{\Vvect}_{1:J}) = - \frac{1}{2} \sum_{j=1}^J \log P (\rvect_j; \thetavect),
\end{equation}
where $\sigma$, $\Qmat$ and $\dvect$ parameterize the rigid transformation that aligns the two point sets. The residual $\rvect_j$ is given by:
\begin{equation}
\rvect_j = \Yvect_j - (\sigma \Qmat (\overline{\Vvect}_j + \Wmat_j \svect) + \dvect), \quad \forall j \in \{1 \dots J\}.
\end{equation}
As above, one can use the generalized Student's t-distribution \eqref{eq:generalized-student} and an expectation-maximization algorithm to robustly estimate the model parameters. The notable difference between \eqref{eq:neg-log-shape} and \eqref{eq:neg-log-L} is that, in addition to scale, rotation and translation, the deformable shape parameters must be estimated as well. The shape parameters are estimated with (please consult Appendix~\ref{app:robust-shape-fit}):
\begin{align}
\label{eq:opt-shape-robust}
\svect = \left( \sum_{j=1}^J \overline{w}_j \Amat_j\tp \Sigmamat\inverse  \Amat_j + \kappa \Lambdamat\inverse\right )\inverse \left( \sum_{j=1}^J \overline{w}_j \Amat_j\tp \Sigmamat\inverse \bvect_j \right),
\end{align}
where $\Amat_j = \sigma \Qmat \Wmat_j$ and $\bvect_j = \Yvect_j - \sigma \Qmat \overline{\Vvect}_j - \dvect$. The shape vertices can now be mapped onto the frontal view, namely $\tilde{\Vvect}_n=(\tilde{V}_{n1}, \tilde{V}_{n2}, \tilde{V}_{n3})$, with:
\begin{equation}
\label{eq:frontal-vertices}
\tilde{\Vvect}_n =  \sigma \Qmat (\overline{\Vvect}_n + \Wmat_n \svect) + \dvect, \quad \forall n \in \{1 \dots N\}.
\end{equation}
A frontal dense depth map is then computed in the following way. 
Remember that the shape vertices form a triangulated 3D mesh; therefore the projection of $\tilde{\Vvect}_{1:N}$ onto the image plane $I_f$ form a 2D triangulated mesh whose vertices have $(\tilde{V}_{n1}, \tilde{V}_{n2})_{1:N}$ as 2D coordinates. Let $n_1$, $n_2$ and $n_3$ be the indexes of the vertices of a mesh triangle. Moreover, let $a_1$, $a_2$ and $a_3$ be the barycentric coordinates of a pixel $(A_1, A_2)$ that lies inside that triangle, with $0 \leq a_1, a_2, a_3 \leq 1$ and $a_1+a_2+a_3=1$. The barycentric coordinates of pixel $(A_1, A_2)$ are computed with:
\begin{equation}
\label{eq:barycenter}
 \begin{pmatrix}
 A_1 \\ A_2
 \end{pmatrix}  = a_1 \begin{pmatrix} \tilde{V}_{n_11} \\ \tilde{V}_{n_12} \end{pmatrix} + a_2 \begin{pmatrix} \tilde{V}_{n_21}\\ \tilde{V}_{n_22}\end{pmatrix} + a_3 \begin{pmatrix} \tilde{V}_{n_31} \\ \tilde{V}_{n_32}\end{pmatrix}.
\end{equation}
Once the barycentric coordinates are thus determined, the depth $A_3$ of the pixel is computed by interpolation, as follows:
\begin{equation}
\label{eq:weighted-interpolation}
A_3 = a_1\tilde{V}_{n_13} + a_2\tilde{V}_{n_23} + a_3\tilde{V}_{n_33}
\end{equation}
The above procedure is repeated for all the triangles and for all the points inside each triangle, thus obtaining a frontal dense depth map of the face.

The final face frontalization step consists of \textit{warping} the input face onto a frontal face.
The rigid transformation from the frontal view to the profile view is the inverse of the pose, namely the inverse of \eqref{eq:landmark-frontalization}: $\rho\pri = \rho\inverse$, $\Rmat\pri = \Rmat\tp$, and $\tvect\pri = -\rho\inverse\Rmat\tp \tvect$. Assuming scaled orthographic projection, a one-to-one correspondence between a pixel  $(A_1,A_2) \in I_f$ with depth $A_3$, and a pixel $(B_1,B_2) \in I_p$ is obtained with the following formula:
\begin{align}
\label{eq:pixel-pred}
\begin{pmatrix}
B_{1} \\ B_{2}
\end{pmatrix} =
\rho\pri 
\begin{pmatrix}
R_{11}\pri & \pri R_{12}\pri & \pri R_{13}\pri \\
R_{21}\pri & \pri R_{32}\pri & \pri R_{33}\pri
\end{pmatrix}
\begin{pmatrix}
A_{1} \\ A_{2}  \\ A_{3} 
\end{pmatrix}
+
\begin{pmatrix}
 t_1\pri \\  t_2\pri
\end{pmatrix}
\end{align}
Finally, face frontalization consists of building an image whose intensities (or colors) are provided by:
\begin{equation}
\label{eq:face-warping}
I_f (A_1, A_2) = I_p ([B_1], [B_2]).
\end{equation}
where $\lbrack \cdot \rbrack$ returns the integer part of a real number.

\section{Algorithm Analysis and Implementation Details}
\label{sec:implementation}
% Implementation

The proposed method is summarized in Algorithm~\ref{algo:face-frontalization}. The method starts with 3D facial landmark extraction which is achieved with \cite{bulat2017far} as it is one of the best-performing methods \cite{sadeghi2020unsupervised}. Moreover, the pose estimation and model fitting steps of 
Algorithm~\ref{algo:face-frontalization} are achieved with Algorithm~\ref{algo:em-student} (Appendix~\ref{app:robust-alignment}) and Algorithm~\ref{algo:em-student-shape}  (Appendix~\ref{app:robust-shape-fit}), respectively. Both these two algorithms are EM procedures and hence they have good convergence properties. 

All the computations inside these two algorithms are in closed-form, with the notable exception of the estimation of the rotation matrix. The latter is parameterized with a unit quaternion \cite{horn1987closed}, which allows us to reduce the number of parameters, from nine to four, and to express the orthogonality constraints of the rotation matrix in a much simpler way. The minimization \eqref{eq:opt-rotation-robust} is solved using a sequential quadratic problem \cite{bonnans2006numerical}. More precisely, a sequential least squares programming (SLSQP) solver\footnote{\url{https://docs.scipy.org/doc/scipy/reference/optimize.html}}
is used in combination with a root-finding software package \cite{kraft1988software}.  The SLSQP minimizer found at the previous EM iteration is used to initialize the current EM iteration. At the start of EM, the closed-form method of
\cite{horn1987closed} is used to initialize the rotation.

%The proposed method also requires a deformable shape model, i.e. Appendix~\ref{app:statistical-shape}. This is implemented in the following way. We consider a set $\mathcal{S}^I= \{\Svect_1^I, \dots, \Svect_m^I, \dots, \Svect_M^I\}$ of $M$ shapes, where each shape $m$ corresponds to a different face identity, and where each face was scanned in a frontal view and with a neutral expression. We also consider a set $\mathcal{S}^E= \{\Svect_1^E, \dots, \Svect_m^E, \dots, \Svect_M^E\}$ that contains $M$ scans of the same faces but with facial expressions, associated one-to-one with the $M$ identities. These two sets are combined in the following way. Let $\Svect_m^\triangle = \Svect_m^I - \Svect_m^E$ be the neutral-expressive difference of face identity $m$. A face $\Svect$ can be reconstructed in the following way:
%
The proposed method also requires a deformable shape model, i.e. Appendix~\ref{app:statistical-shape}. We consider a set $\mathcal{S}^I= \{\Svect_1^I, \dots, \Svect_m^I, \dots, \Svect_M^I\}\subset \mathbb{R}^{3N}$ of $M$ shapes, where $\Svect_m^I$ is a concatenation of $N$ 3D coordinates that corresponds to a different face identity, and where each face was scanned in a frontal view and with a neutral expression. We also consider a set $\mathcal{S}^E= \{\Svect_1^E, \dots, \Svect_m^E, \dots, \Svect_M^E\}\subset \mathbb{R}^{3N}$ that contains $M$ scans of the same faces but with facial expressions, associated one-to-one with the $M$ identities. Let $\Svect_m^\triangle = \Svect_m^E - \Svect_m^I$ be the expressive-neutral difference of face identity $m$, namely the expressive offset. 
%It is intuitive that a linear combination of the identity set together with a linear combination of the expressive offset set can model an unseen face with expression. 
A face $\Svect$ can be reconstructed from its identity and expression embeddings, i.e. Appendix~\ref{app:statistical-shape}:
\begin{align}
\label{eq:shape-embedding-practical}
\hat{\Svect} = \overline{\Svect}^I + \overline{\Svect}^\triangle + \widetilde{\Umat}^I \svect^I +  \widetilde{\Umat}^\triangle\svect^\triangle,
\end{align}
where $\overline{\Svect}^I$ and $\overline{\Svect}^\triangle$ are the means  associated with the identity- and with the expressive-neutral difference, matrices $\widetilde{\Umat}^I$ and $\widetilde{\Umat}^\triangle$ contain the $K$ principle eigenvectors, and $\svect^I$ and $\svect^\triangle$ are the corresponding embeddings. Note that we use $\hat{\Svect}$ to emphasize that the reconstruction is an approximation of $\Svect$.

%where $\svect^I$ and $\svect^\triangle$ are the shape embeddings associated with the identity and with the neutral-expressive difference, respectively.
%

In practice we use the publicly available Basel Shape Model (BSM) \cite{paysan20093d} augmented with \cite{cao2013facewarehouse}. 
This provides $M=200$ registered face scans in a frontal view and with neutral expressions, corresponding to $M$ different identities, as well as $M$ expressive scans of the same identities. Each scan is described by a triangulated mesh with an identical number of vertices, namely $N=53490$. The dimension of the embedding is $K=200$ and hence we have $K\ll 3N$. We use the landmark locations associated with the mean identity $ \overline{\Svect}^I$ to compute the pose, i.e $\Zvect_{1:J}$.

\section{Experiments}
\label{sec:experiments}
% Experiments

%\begin{itemize}
%\item  Empirical validation based on correlation between frontalized face and ground-truth frontal view of the same face [DONE] 
%\item  Describe the dataset used for validation [DONE]
%\item  Frontalization benchmark and results [DONE]
%\item Additional illustrative experiments
%\item Show failure cases and analyse them
%\end{itemize}

As with any methodological development, performance evaluation is extremely important. In this paper we make recourse to empirical evaluation based on a dataset with associated ground truth. Such a dataset should contain pairs of frontal and profile views of faces for a large number of subjects. Quantitative performance evaluation consists of computing a metric between an image obtained by face frontalization of a profile view of a subject, with an image containing a frontally-viewed face of the same subject. It is desirable that the profile and frontal images are simultaneously recorded with two synchronized cameras. Therefore, the proposed evaluation is based on image-to-image comparison. Several metrics were developed in the past for comparing two images, e.g. feature-based and pixel-based metrics. In this work we use the \textit{zero-mean normalized cross correlation} (ZNCC) coefficient between two image regions, a measure that has successfully been used for stereo matching, e.g. \cite{sun2002fast}, because it is invariant to differences in brightness and contrast between the two images, due to the normalization with respect to mean and standard deviation. 

Let $R_f (h,v)\subset I_f$ be a region of size $H\times V$ whose center coincides with pixel location $(h,v)$ of a frontalized image $I_f$. Similarly, let $R_t (h,v)\subset I_t$ be a region of the same size and whose center coincides with pixel location $(h,v)$ of a ground-truth image $I_t$. The ZNCC coefficient between these two regions writes:
\begin{align}
\label{eq:zncc}
C (h,v, & h\pri, v\pri) = \\
  \max_{\delta h, \delta v} & \bigg\{ \frac{\cov [R_f (h,v),R_t (h+\delta h, v+ \delta v)]}{\var [R_f (h,v)]^{1/2} \var [R_t (h+\delta h,v+\delta v)]^{1/2}  } \bigg\}, \nonumber
\end{align}
where $\cov[\cdot, \cdot]$ is the centered covariance between the two regions, $\var [\cdot]$ is the centered variance of a region, $\delta h$ and $\delta v$ are horizontal and vertical shifts, and $h\pri$ and $v\pri$ are the horizontal and vertical shifts that maximize the ZNCC coefficient.

In order to evaluate the performance of the proposed frontalization method and to compare it with state-of-the-art methods, we used a publicly available dataset, namely the OulouVS2 dataset \cite{anina2015ouluvs2}. This dataset targets the understanding of speech perception, more precisely, the analysis of non-rigid lip motions that are associated with speech production. The dataset was recorded in an office with ordinary (artificial and natural) lighting conditions.  The recording setup consists of five synchronized cameras (2~MP, 30~FPS) placed in five different points of view, namely $0^{\circ}$, $30^{\circ}$, $45^{\circ}$, $60^{\circ}$ and $90^{\circ}$. 

The dataset contains $5 \times 780$ videos recorded with 53 participants. Each participant was instructed to read loudly several text sequences displayed on a computer monitor placed slightly to the left and behind the $0^{\circ}$ (frontal) camera. The displayed text consists of digit sequences, e.g. ``one, seven, three, zero, two, nine", of phrases, e.g. ``thank you", ``have a good time", and ``you are welcome", as well as of sequences from the TIMIT dataset, e.g. ``agricultural products are unevenly distributed". While participants were asked to keep their head still, natural uncontrolled head movements and body position changes were inevitable. As a consequence the actual head pose varies from one participant to another and there is no exact match between the head and camera orientations. 

   \begin{table}[t!]
    \begin{center}
      \caption{ \label{table:zncc} Mean ZNCC coefficients for 15 participants. The best score is in \textbf{bold} and the second best is in \textit{\textbf{slanted bold}}. }
    \begin{tabular}{ |l|l|c|  }
%     \hline
%     \multicolumn{3}{|c|}{Experiment on OuluVS2 data set} \\
     \hline
     Method & Post-processing & ZNCC\\
     \hline
     Hassner et al. \cite{hassner2015effective}  & -    &0.723\\
     Hassner et al.  \cite{hassner2015effective}  &   Soft symmetry  & 0.780\\
     Hassner et al.  \cite{hassner2015effective}  & Hard symmetry  & 0.722\\
     Banerjee et al. \cite{banerjee2018frontalize} & -  & 0.739\\
     Banerjee et al.  \cite{banerjee2018frontalize} & Soft symmetry & \textit{\textbf{0.788}}   \\
     Proposed & - & \textbf{0.824} \\
     \hline
    \end{tabular}
       \end{center}
    \end{table}
    
    % Table of ZNCC w.r.t. yaw angles
\begin{table}[t!]
\begin{center}
\caption{\label{table:yawangles} Results for nine participants as a function of estimated yaw angle (in degrees) that corresponds to the horizontal head orientation computed with the proposed 3D head-pose estimator. Both \cite{hassner2015effective} and \cite{banerjee2018frontalize} make use of symmetry to fill in the occluded face areas. The best scores are in \textbf{bold} and the second best are in \textit{\textbf{slanted bold}}.}
\begin{tabular}{|c|c|c|c|c|}
 \hline
% \multicolumn{5}{|c|}{ZNCC score with respect to the yaw angles} \\
% \hline
 Participant & Yaw  & \cite{hassner2015effective}  & \cite{banerjee2018frontalize} & Proposed \\
 \hline
 \#31  & 19.1 & \textit{\textbf{0.905}} & 0.856 & \textbf{0.924}\\
 \#01   & 23.5   & \textbf{0.915} & 0.893& \textit{\textbf{0.913}}\\
 \#02  & 24.9   & \textit{\textbf{0.888}} & 0.878 & \textbf{0.948} \\
 \#10 & 29.0 & 0.805 & \textbf{0.812} & \textit{\textbf{0.806}}\\
 \#23 & 30.0 & 0.810 & \textbf{0.857} & \textit{\textbf{0.842}} \\
 \#27 & 32.9 & 0.685 & \textbf{0.852} & \textit{\textbf{0.732}} \\
 \#19 & 37.8 &\textit{\textbf{0.752}} & 0.650 & \textbf{0.754} \\
 \#12 & 38.5& \textit{\textbf{0.731}} & 0.713 & \textbf{0.774} \\
 \#21 & 40.6 & 0.632 & \textbf{0.743} & \textit{\textbf{0.735}} \\
 \hline
 Mean $\&$ std. & & 0.791$\pm$0.100 & \textit{\textbf{0.806}}$\pm$0.084 & \textbf{0.825}$\pm$0.085 \\
 \hline
\end{tabular}
\end{center}
\end{table}

% Figure of Oulu data set    
\begin{figure*}[p!]
   \centering
\begin{tabular}{cccc}
%(a)&
\includegraphics[trim = 14cm 8cm 27cm 6cm,clip,keepaspectratio=true,width=0.22\linewidth]{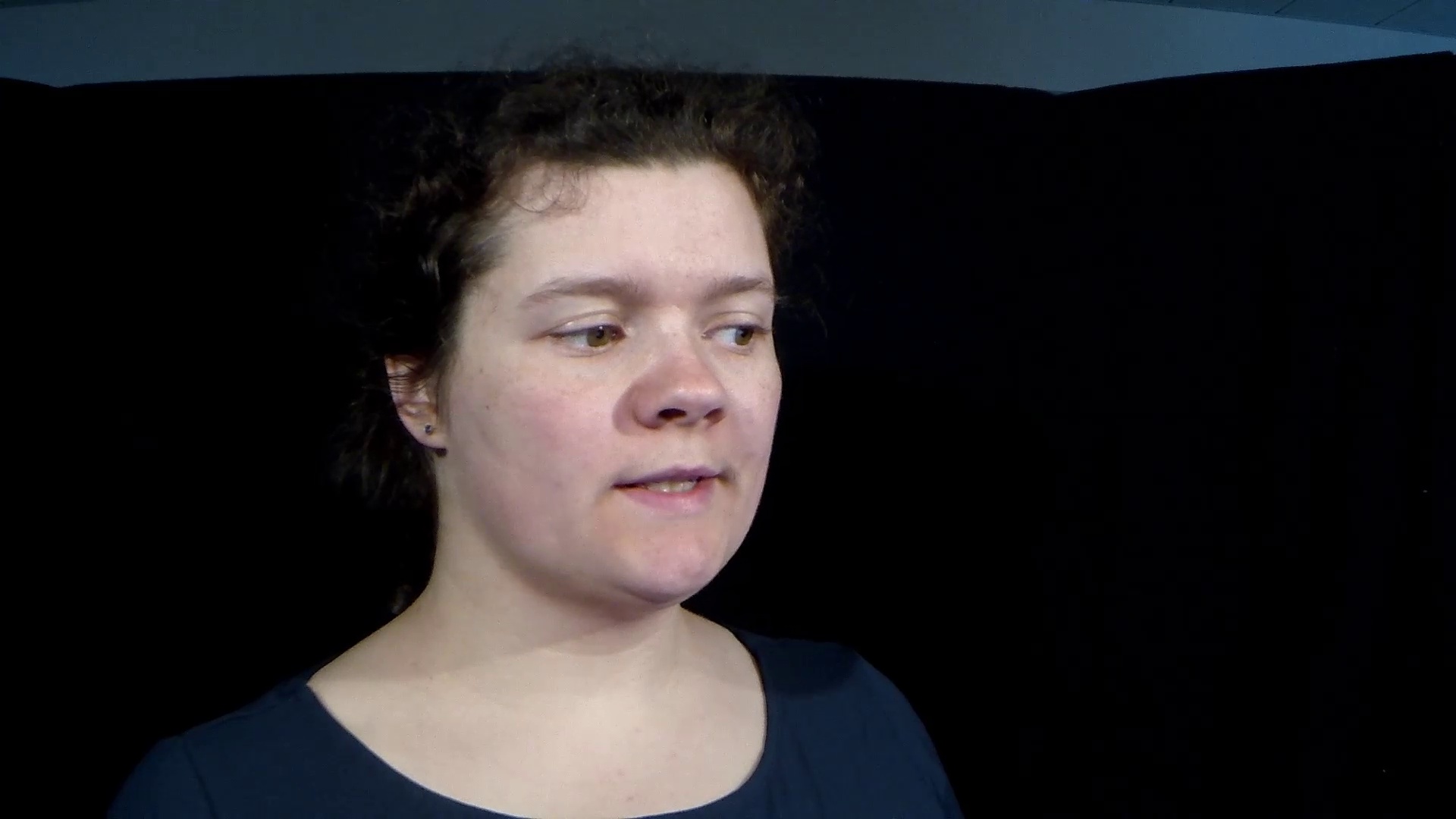}&
\includegraphics[trim = 14cm 8cm 27cm 6cm,clip,keepaspectratio=true,width=0.22\linewidth]{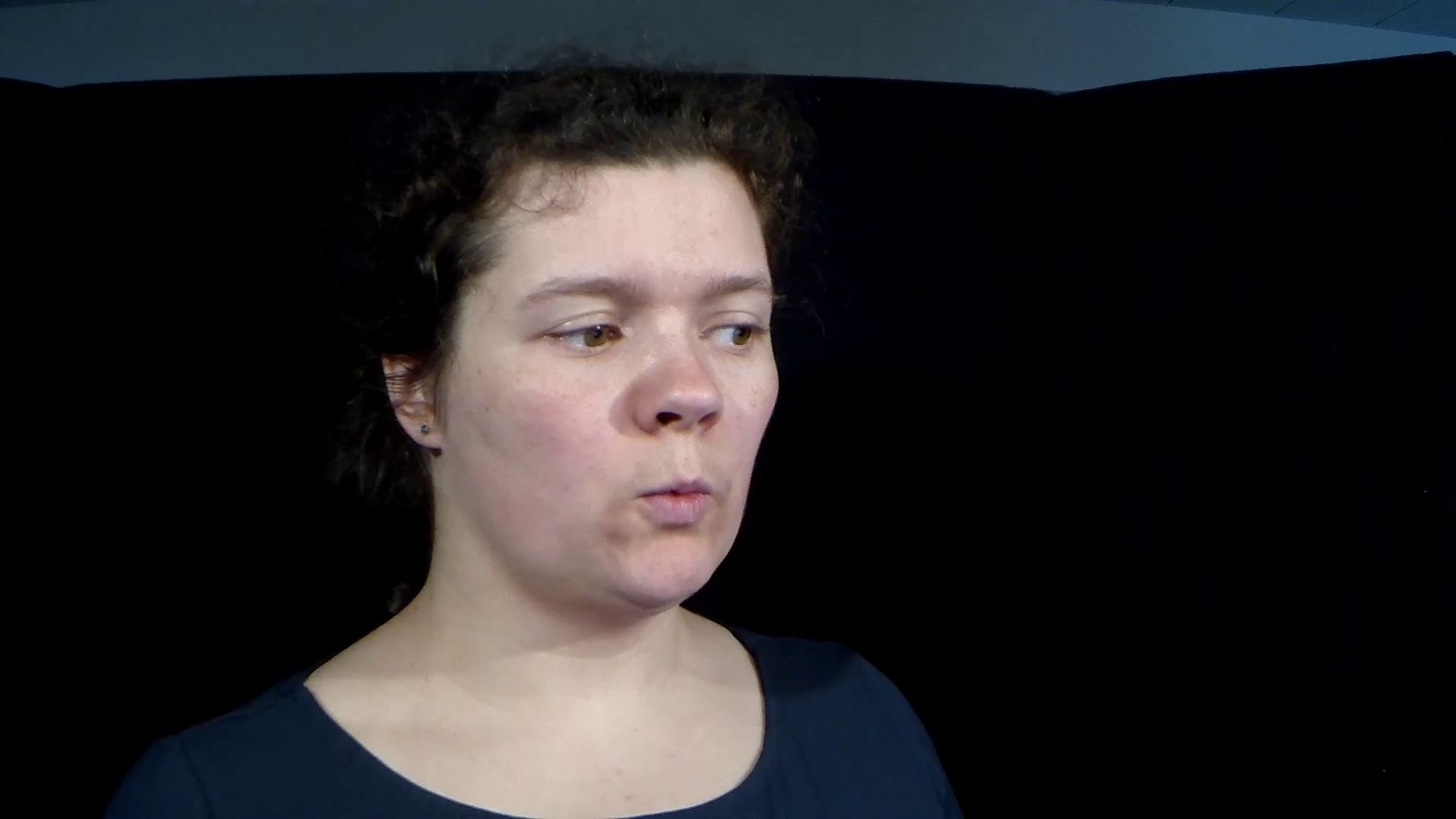}&
\includegraphics[trim = 14cm 8cm 27cm 6cm,clip,keepaspectratio=true,width=0.22\linewidth]{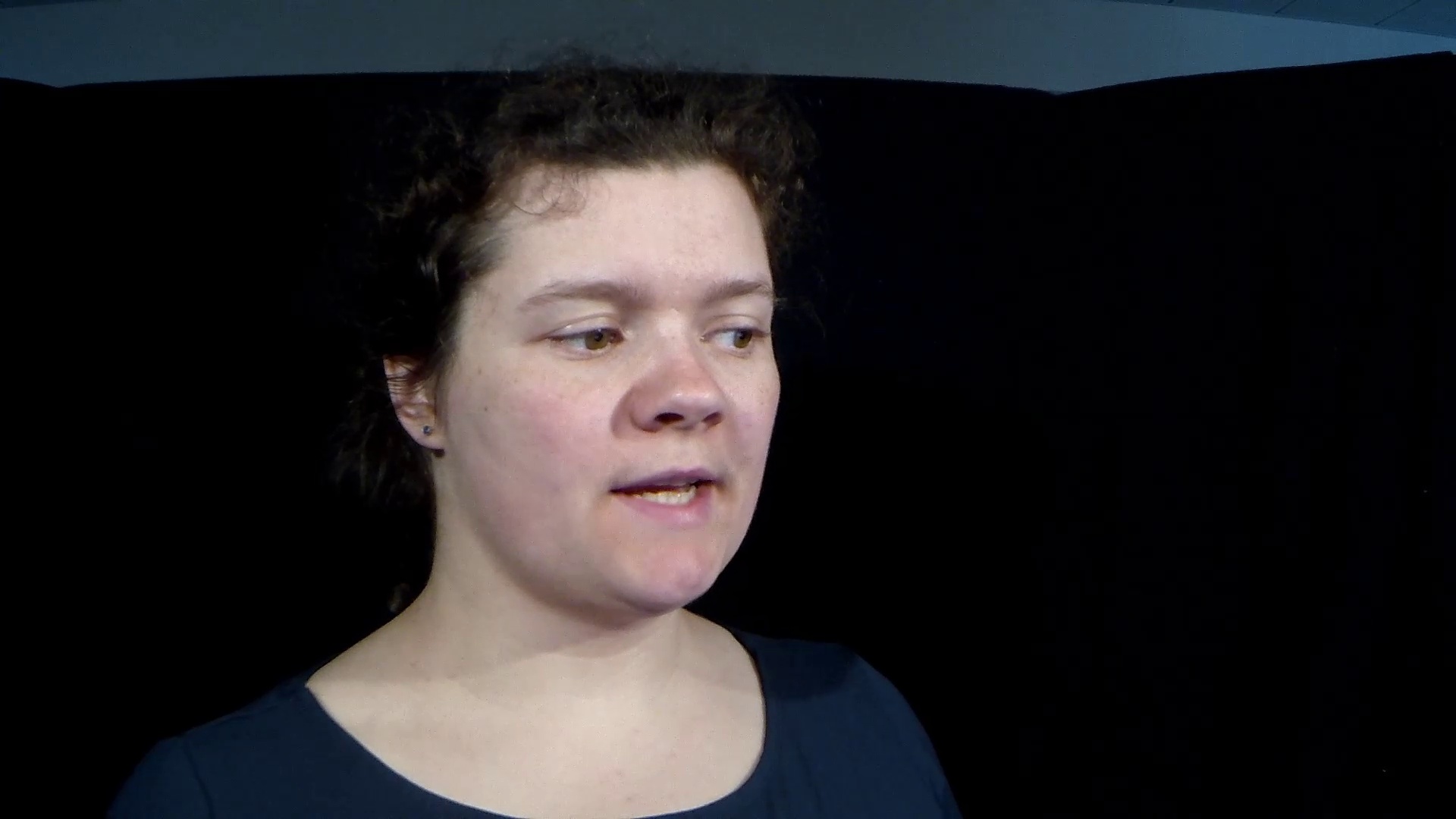}&
\includegraphics[trim = 14cm 8cm 27cm 6cm,clip,keepaspectratio=true,width=0.22\linewidth]{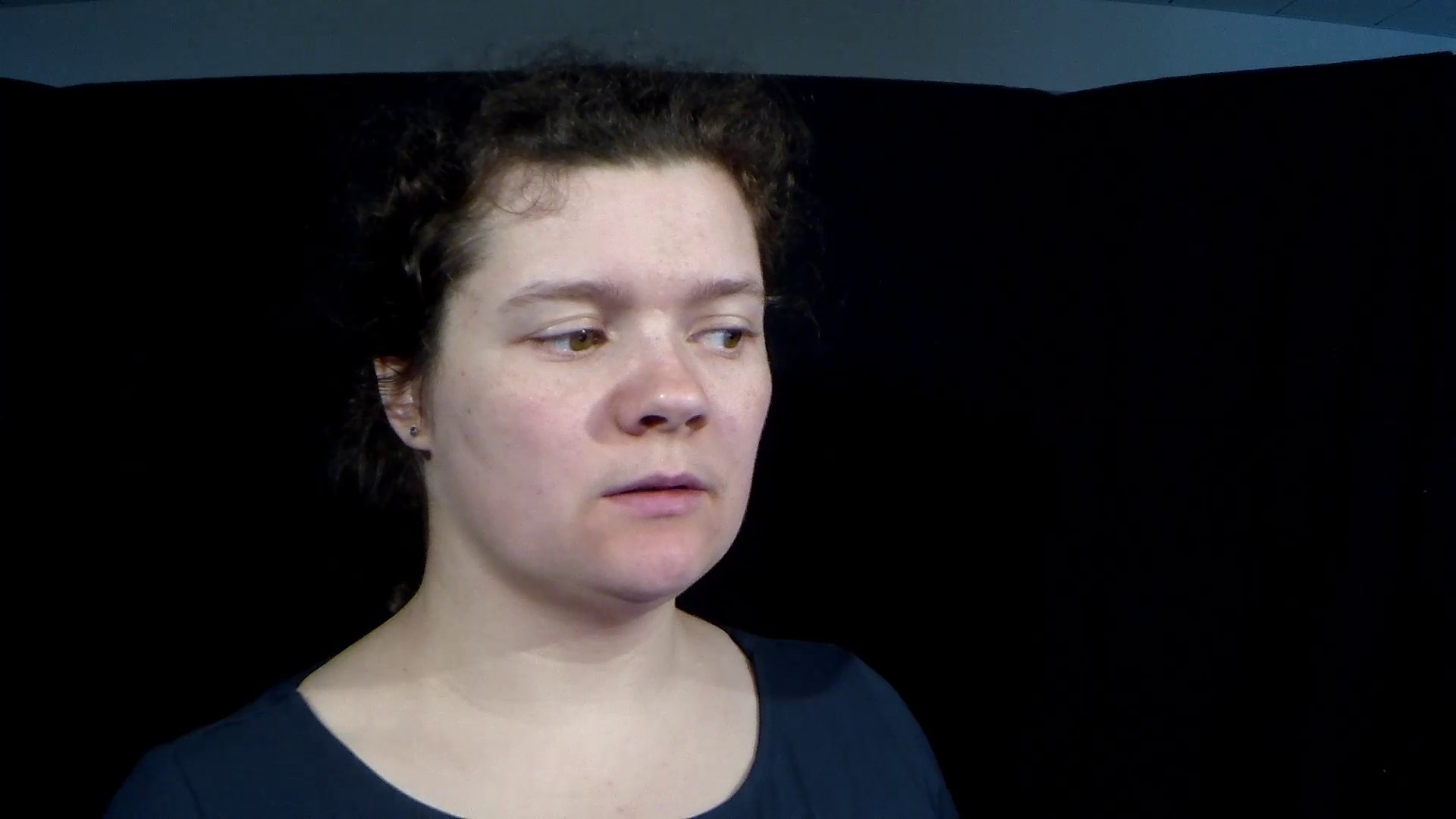}\\
%(b)&
\includegraphics[trim = 20cm 6cm 20cm 8cm,clip,keepaspectratio=true,width=0.22\linewidth]{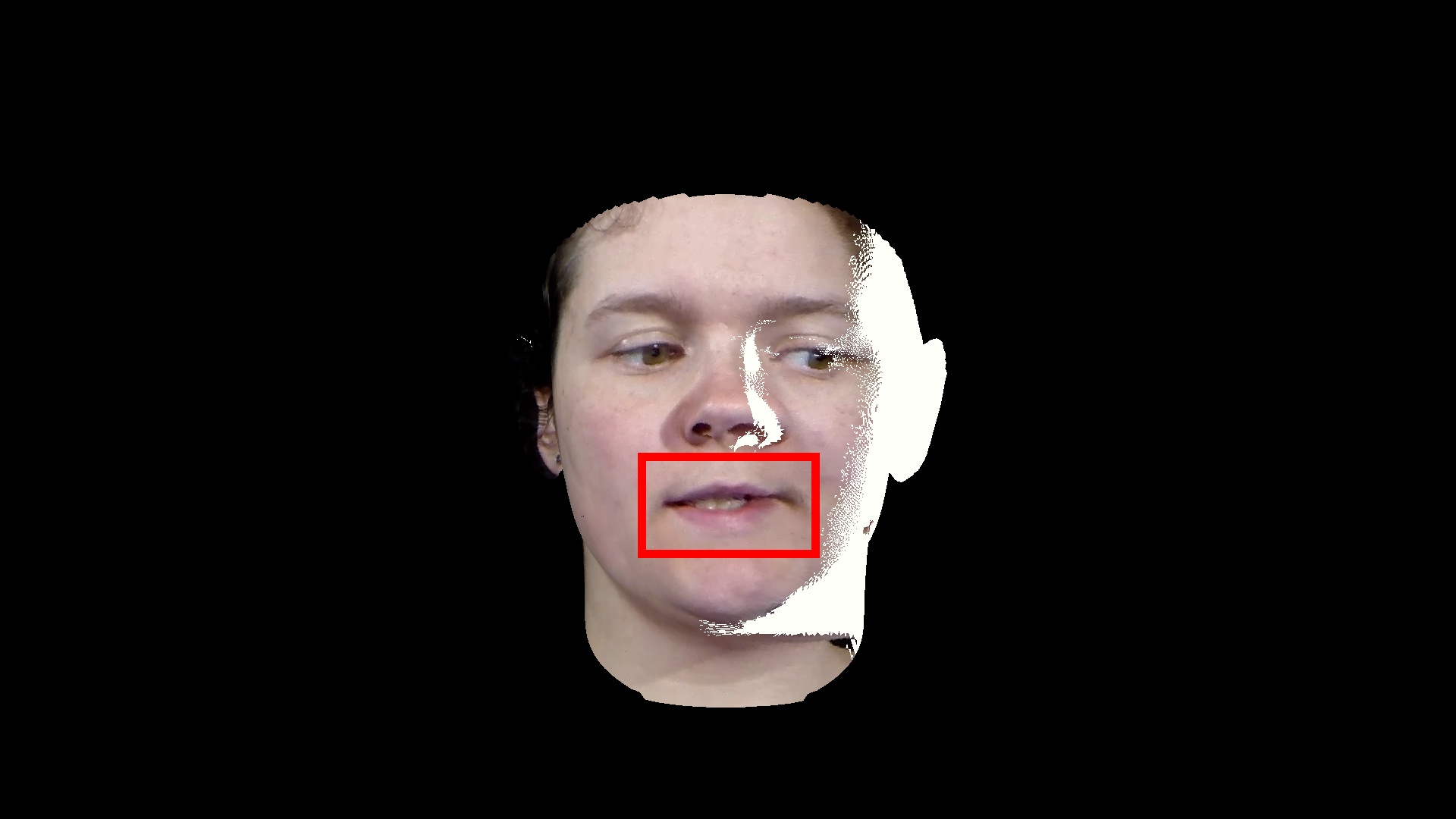}&
\includegraphics[trim = 20cm 6cm 20cm 8cm,clip,keepaspectratio=true,width=0.22\linewidth]{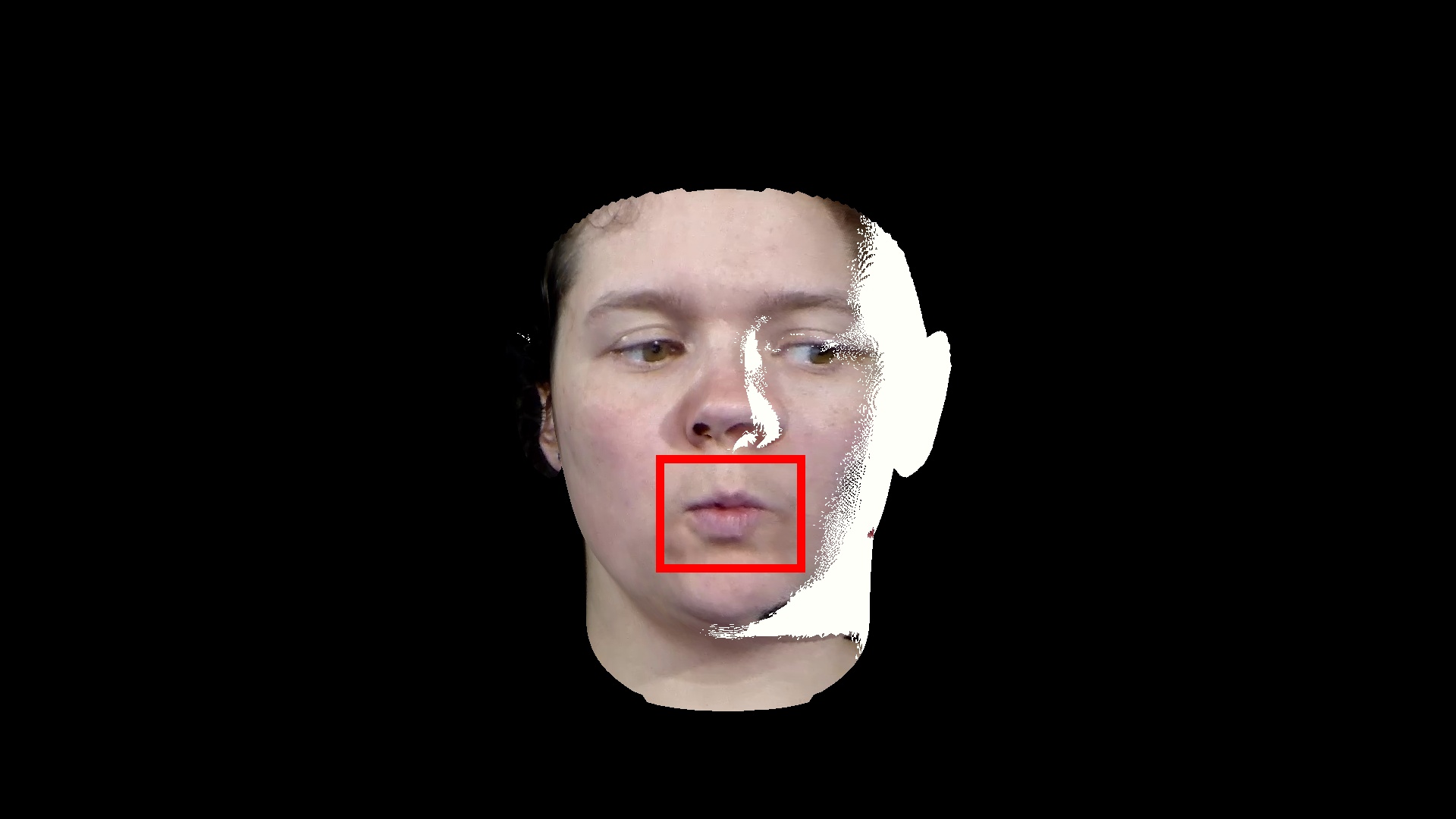}&
\includegraphics[trim = 20cm 6cm 20cm 8cm,clip,keepaspectratio=true,width=0.22\linewidth]{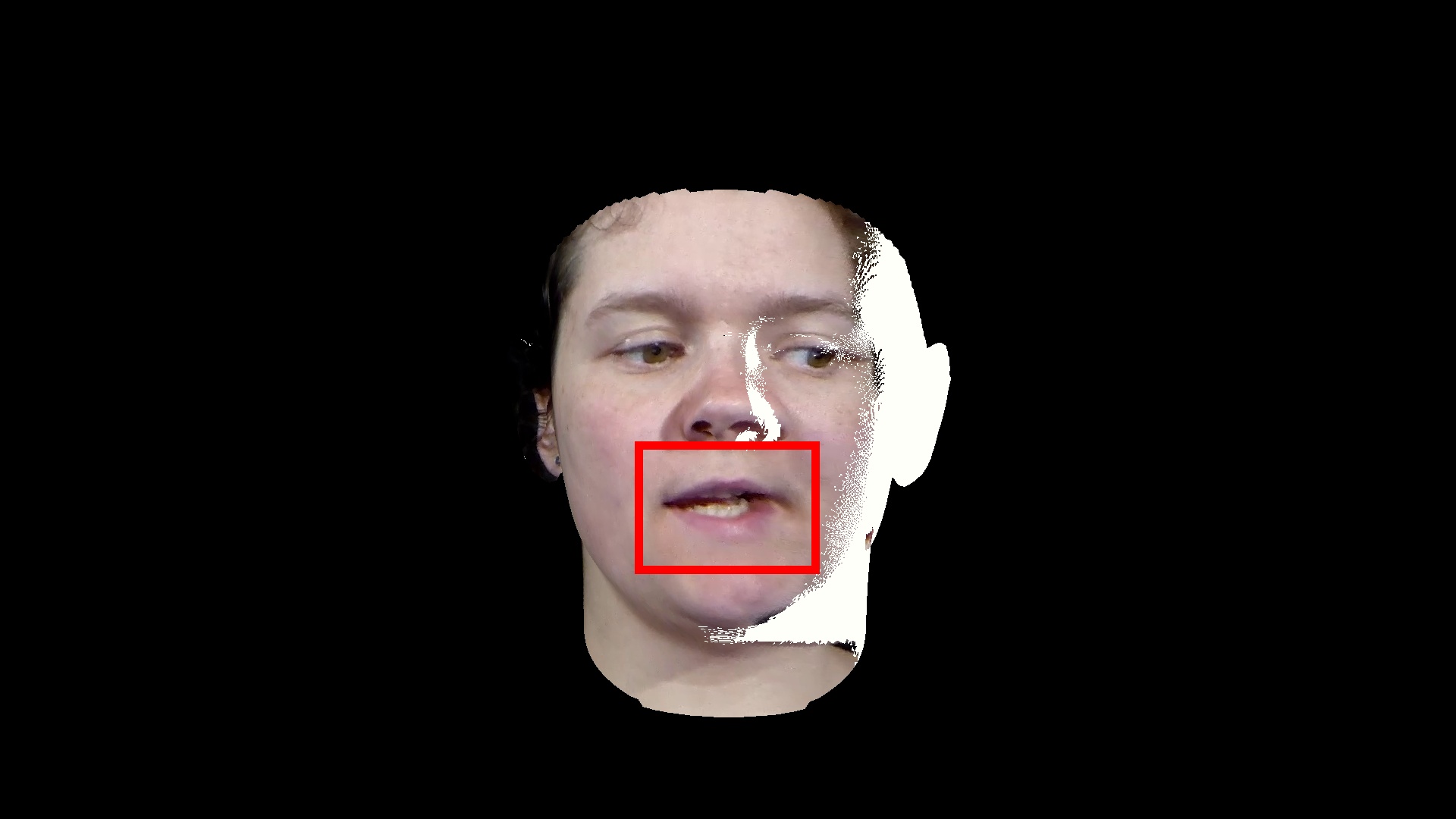}&
\includegraphics[trim = 20cm 6cm 20cm 8cm,clip,keepaspectratio=true,width=0.22\linewidth]{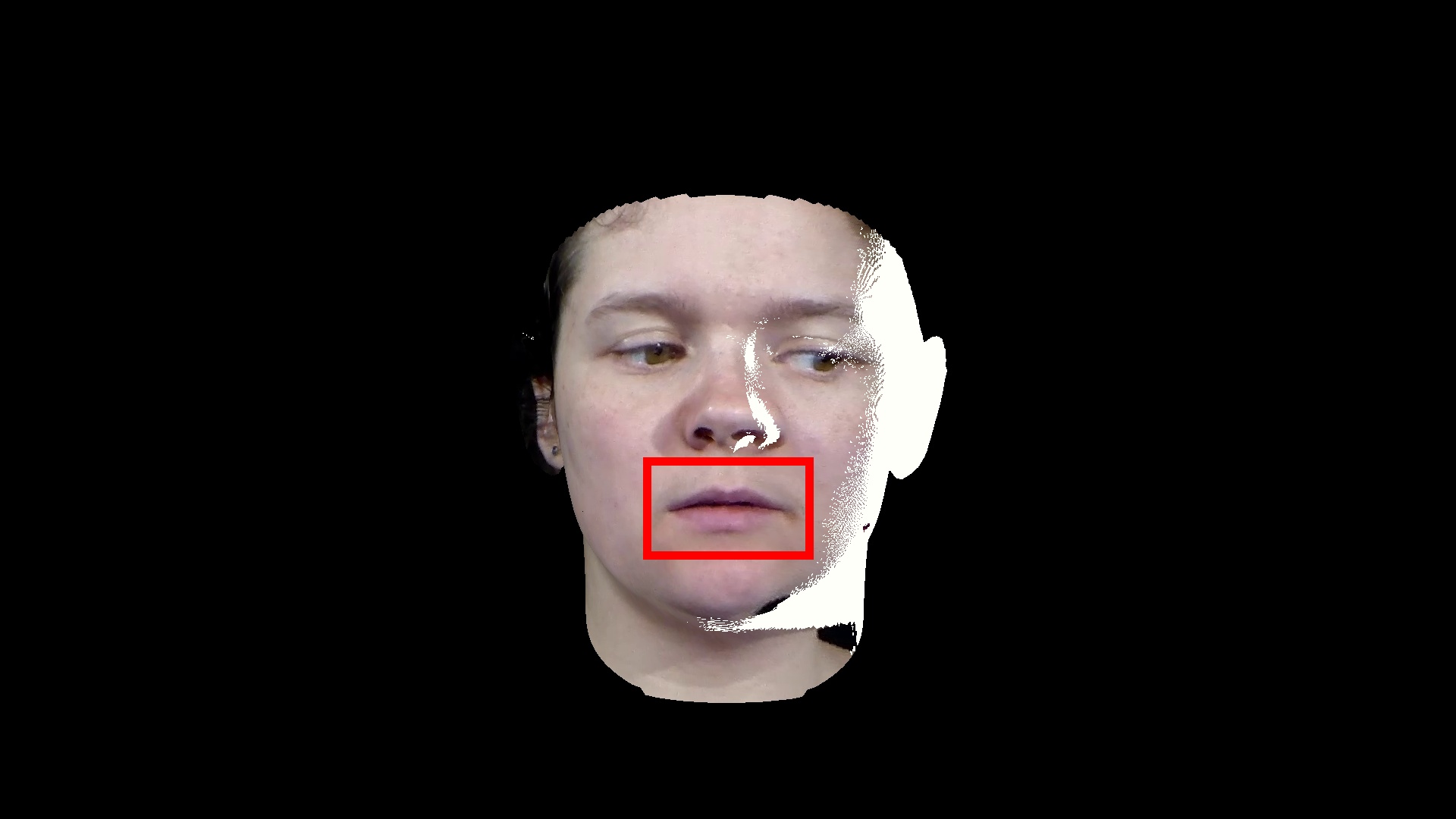}\\
%(c) &
0.940&0.974&0.925&0.972\\
\includegraphics[trim = 22cm 10cm 18cm 4cm,clip,keepaspectratio=true,width=0.22\linewidth]{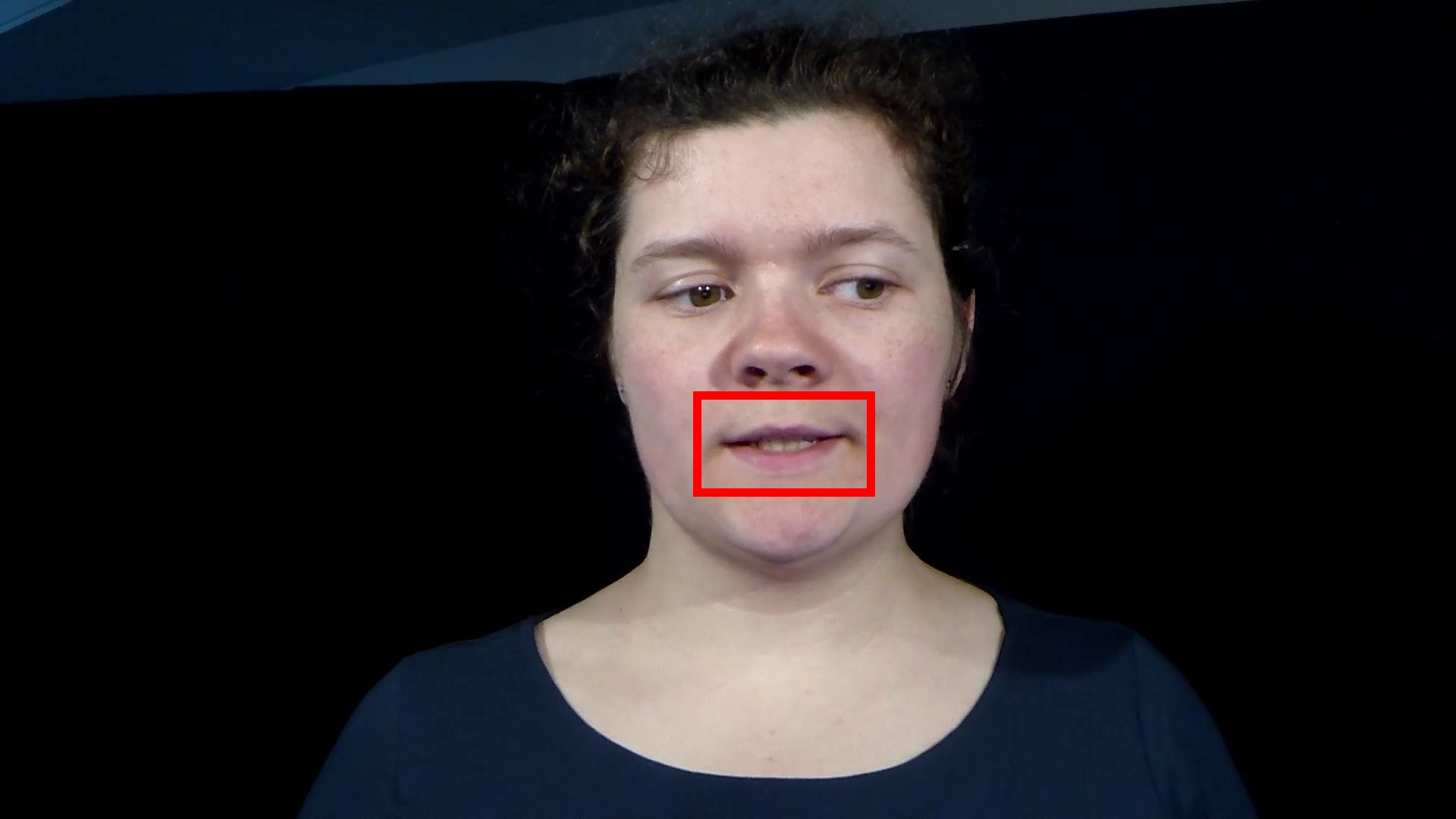}&
\includegraphics[trim = 22cm 10cm 18cm 4cm,clip,keepaspectratio=true,width=0.22\linewidth]{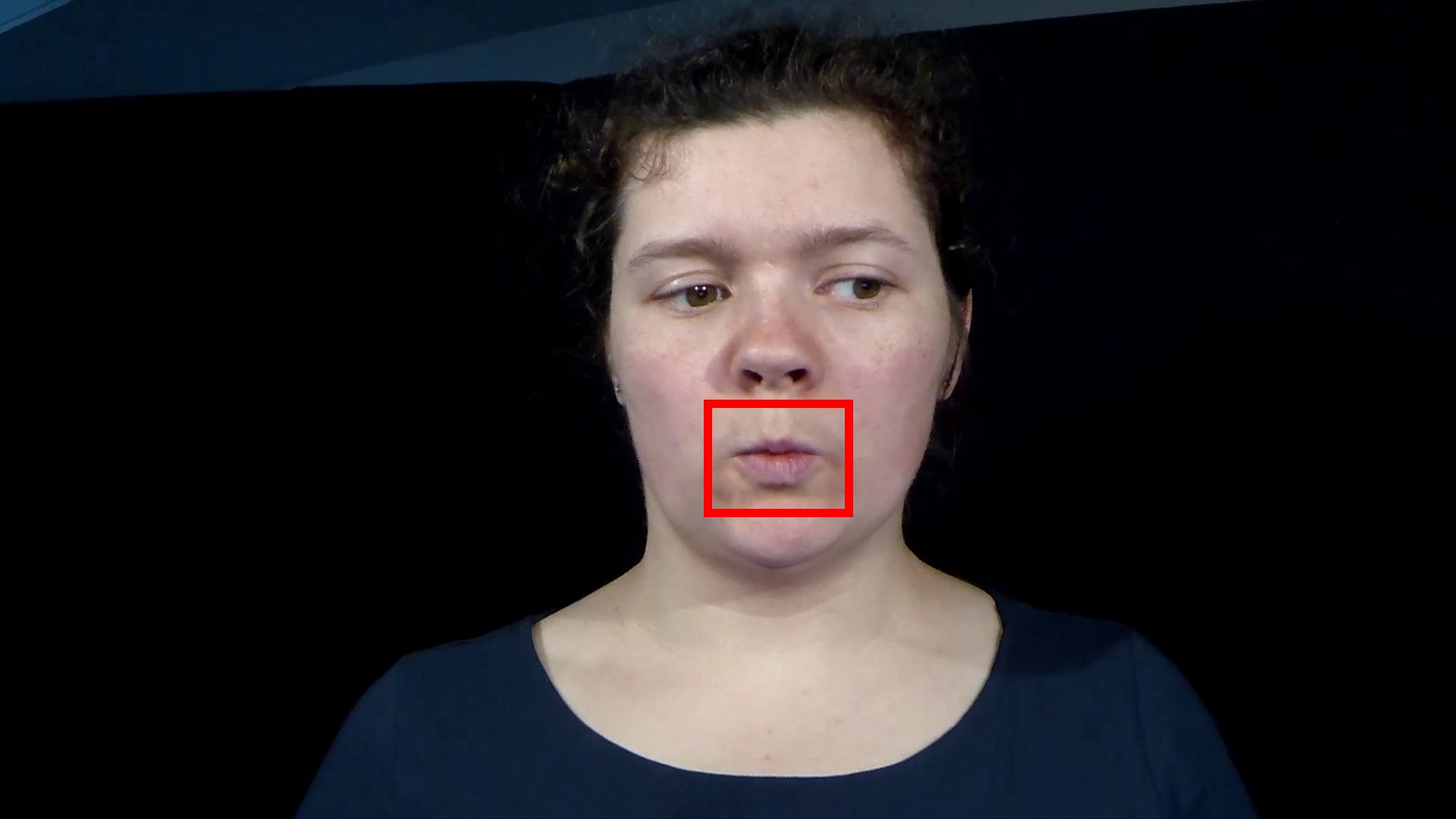}&
\includegraphics[trim = 22cm 10cm 18cm 4cm,clip,keepaspectratio=true,width=0.22\linewidth]{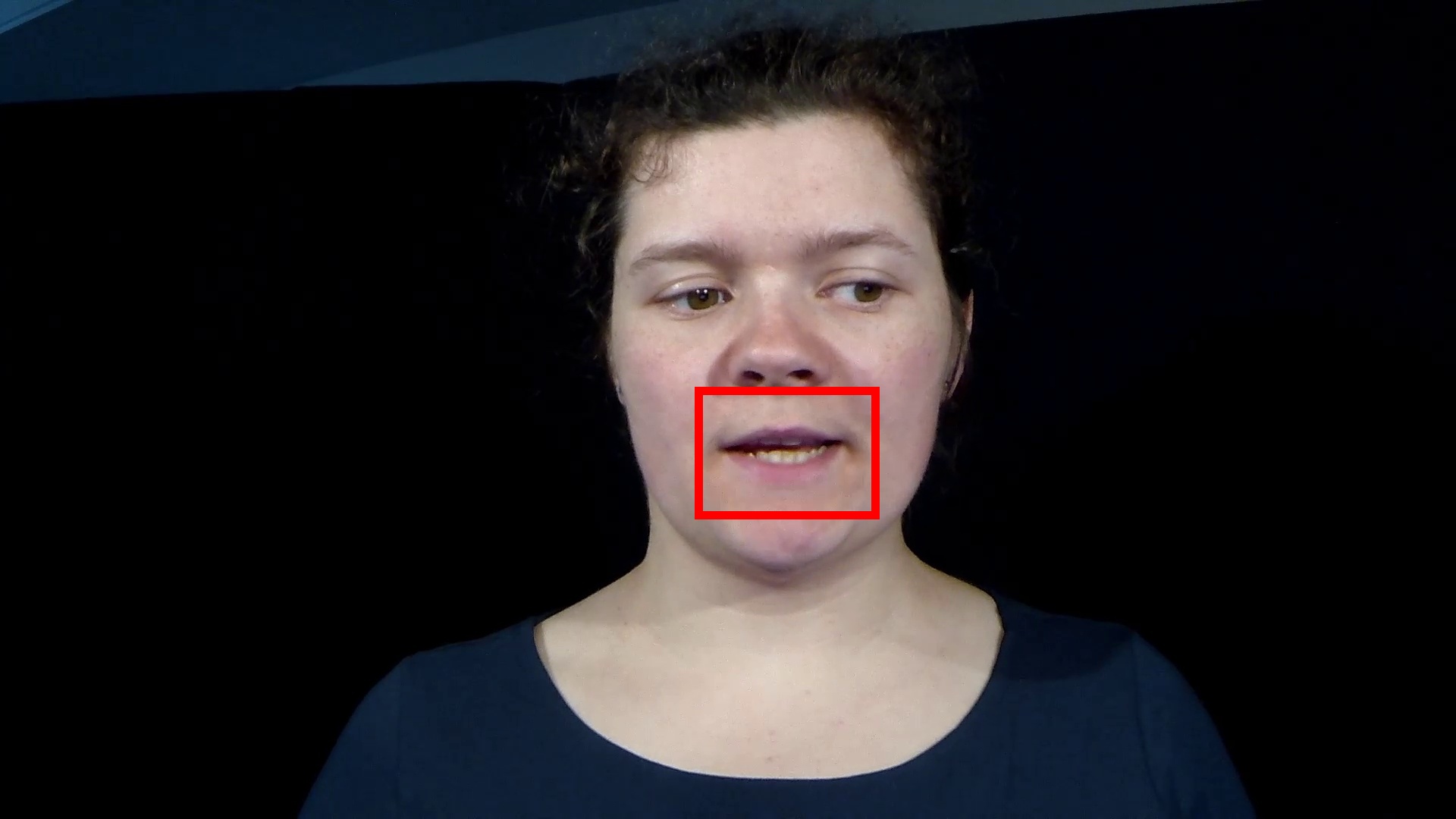}&
\includegraphics[trim = 22cm 10cm 18cm 4cm,clip,keepaspectratio=true,width=0.22\linewidth]{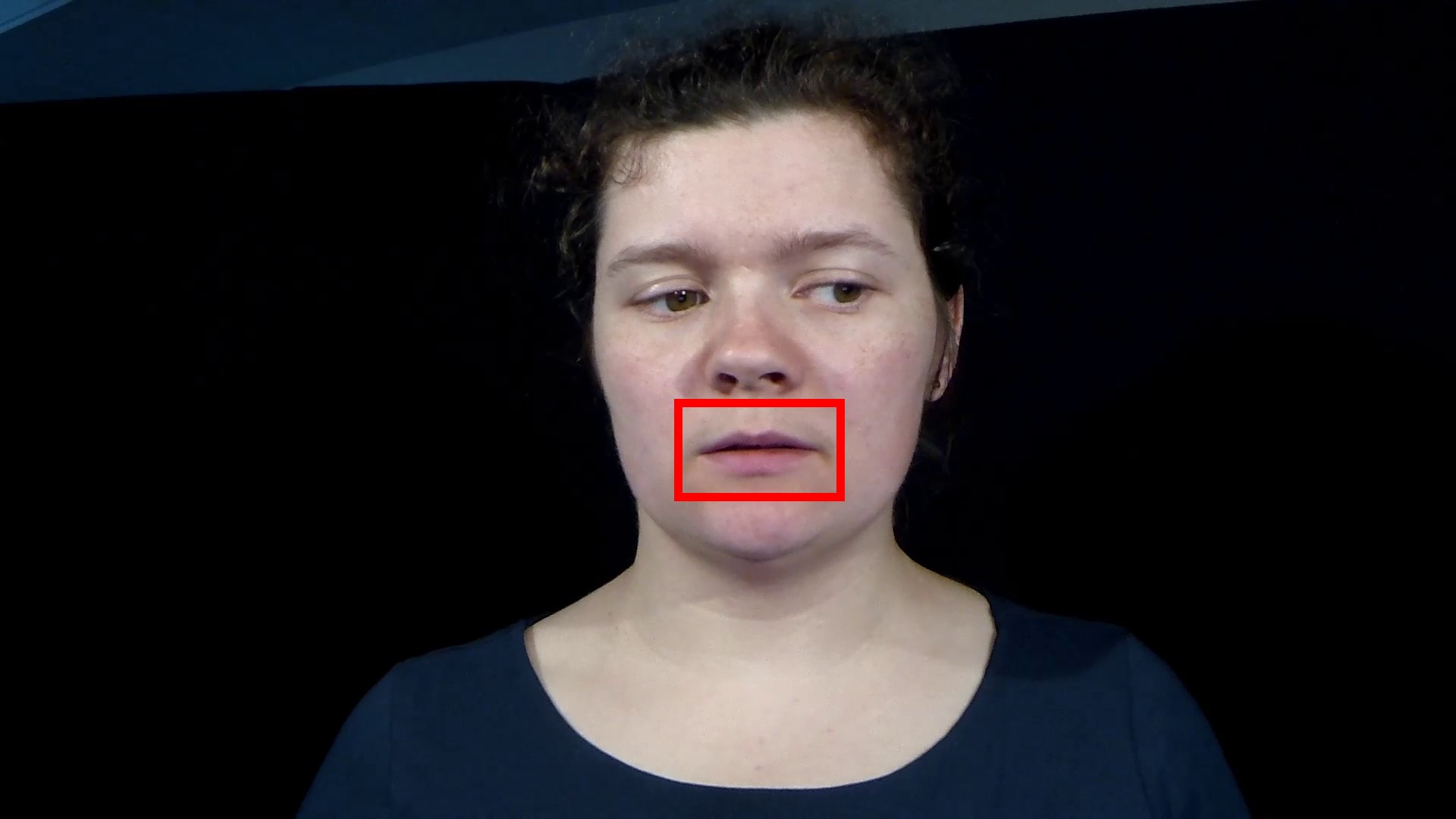}
%(d)&
\end{tabular}
    \caption{Face frontalization exemples and ZNCC coefficients obtained with our method for participant \#02 from the OulouVS2 dataset. Top row: input face recorded with the $30^\circ$ camera. The estimated horizontal head orientation (yaw angle) is of $25^\circ$ in this case. Middle row: frontalization results with self occluded facial regions displayed in white. Bottom row: ground-truth frontal face recorded with the $0^\circ$ camera. The ZNCC coefficients correspond to the mouth bounding boxes shown in red.}
    \label{fig:Oulu2}
\end{figure*}    

\begin{figure*}[p!]
    \centering
\begin{tabular}{cccc}
\includegraphics[width=0.22\linewidth]{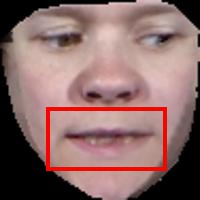}&
\includegraphics[width=0.22\linewidth]{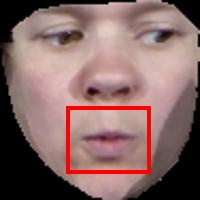}&
\includegraphics[width=0.22\linewidth]{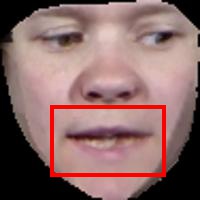}&
\includegraphics[width=0.22\linewidth]{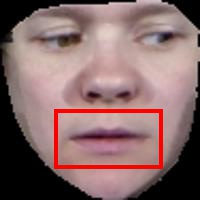} \\
0.856&0.946&0.880&0.961\\
\includegraphics[width=0.22\linewidth]{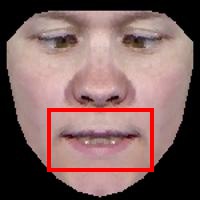}&
\includegraphics[width=0.22\linewidth]{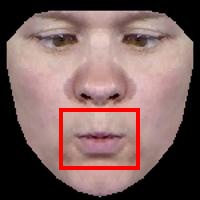}&
\includegraphics[width=0.22\linewidth]{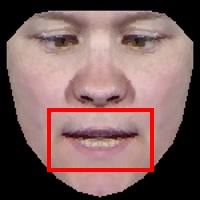}&
\includegraphics[width=0.22\linewidth]{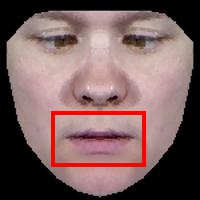}\\
0.819&0.923&0.860&0.929
\end{tabular}
    \caption{Face frontalization results and ZNCC coefficients for participant \#02. Top row: Hassner et al \cite{hassner2015effective}. Bottom row: Banerjee et al. \cite{banerjee2018frontalize}. Both these methods enforce symmetry to fill in the gaps caused by self occlusions. }
    \label{fig:Oulu_Hassner_Banerjee_2}
\end{figure*}

\begin{figure*}[p!]
   \centering
\begin{tabular}{cccc}
%(a)&
\includegraphics[trim = 9cm 1cm 27cm 6cm,clip,keepaspectratio=true,width=0.22\linewidth]{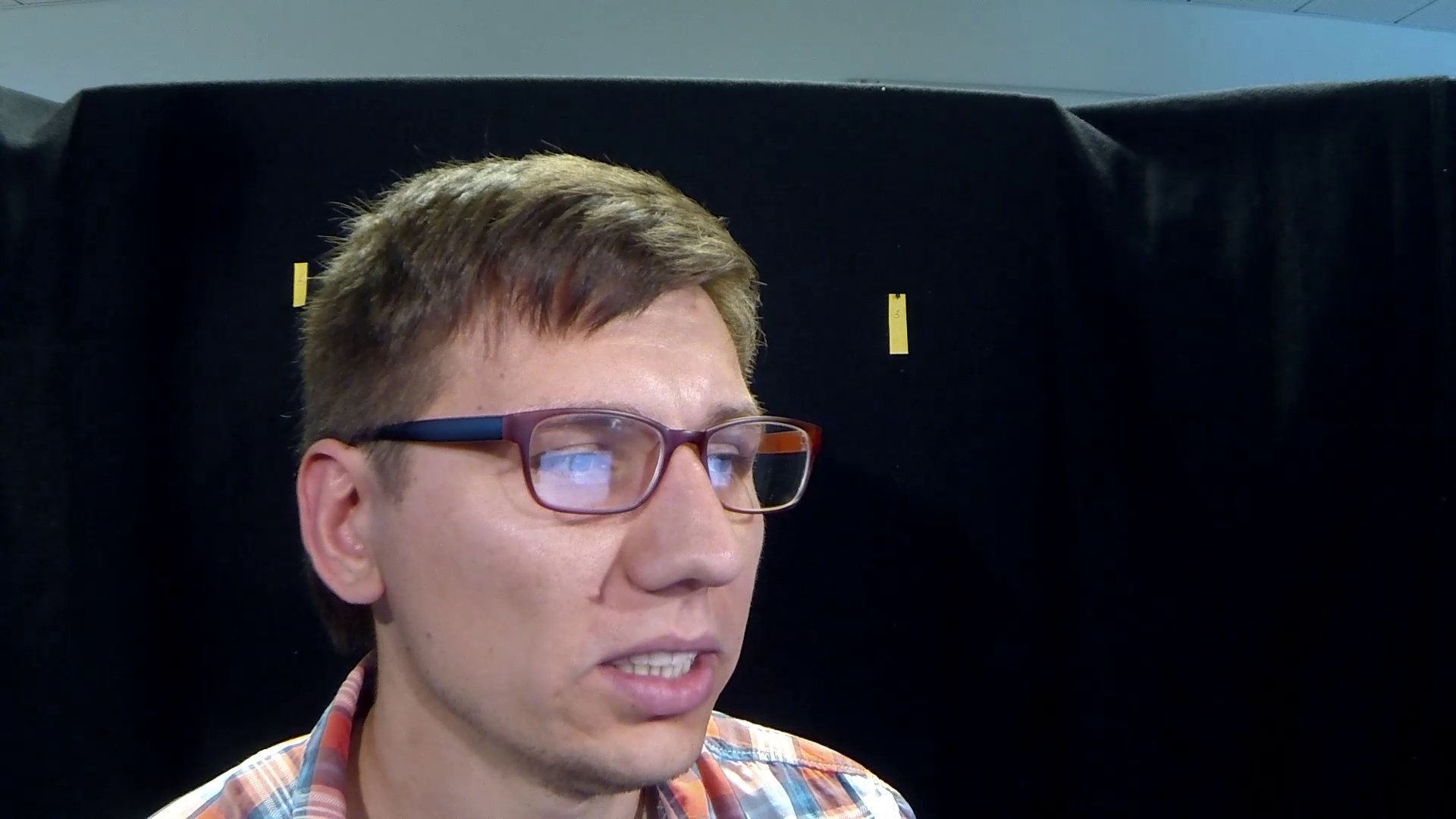}&
\includegraphics[trim = 9cm 1cm 27cm 6cm,clip,keepaspectratio=true,width=0.22\linewidth]{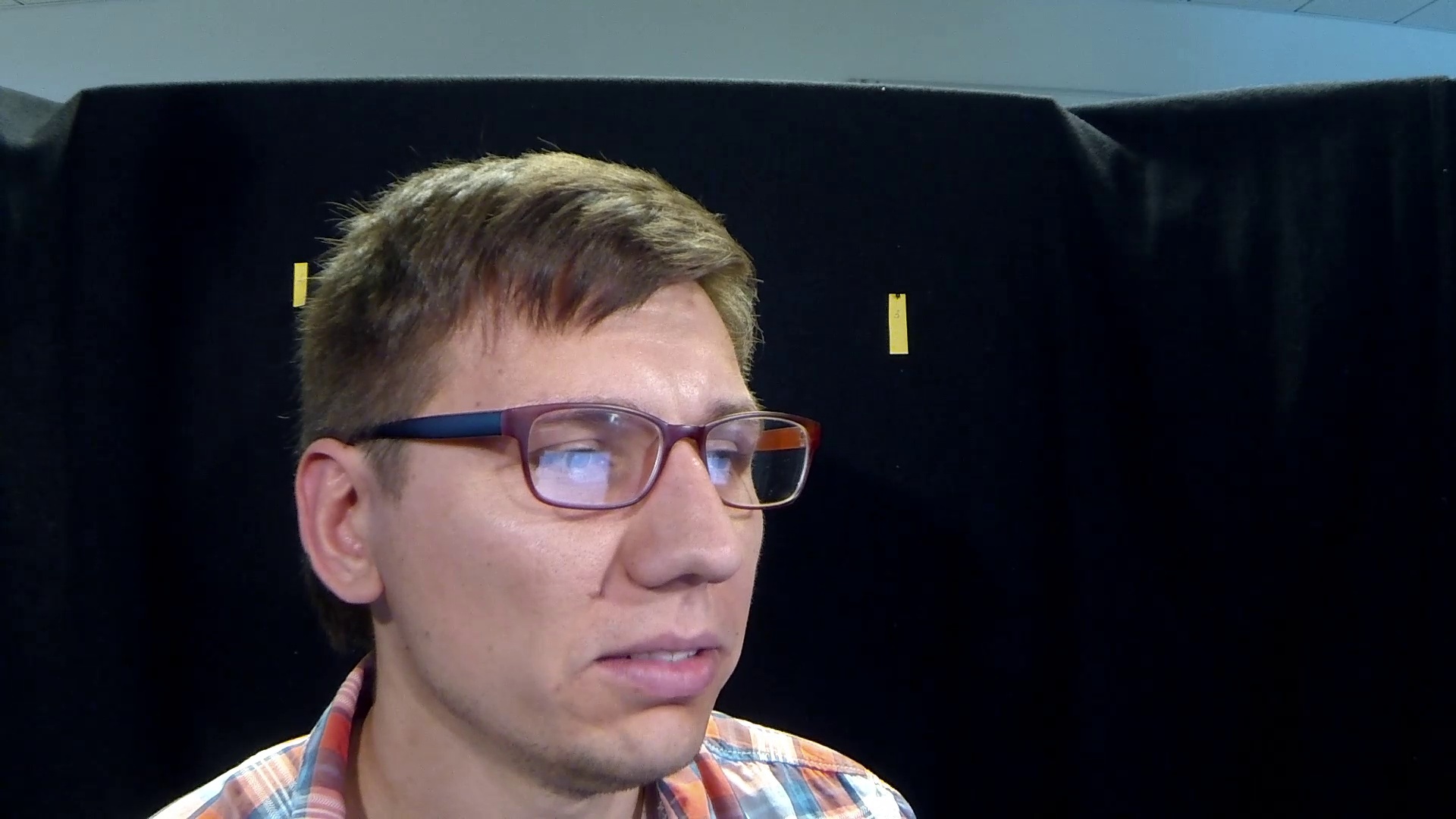}&
\includegraphics[trim = 9cm 1cm 27cm 6cm,clip,keepaspectratio=true,width=0.22\linewidth]{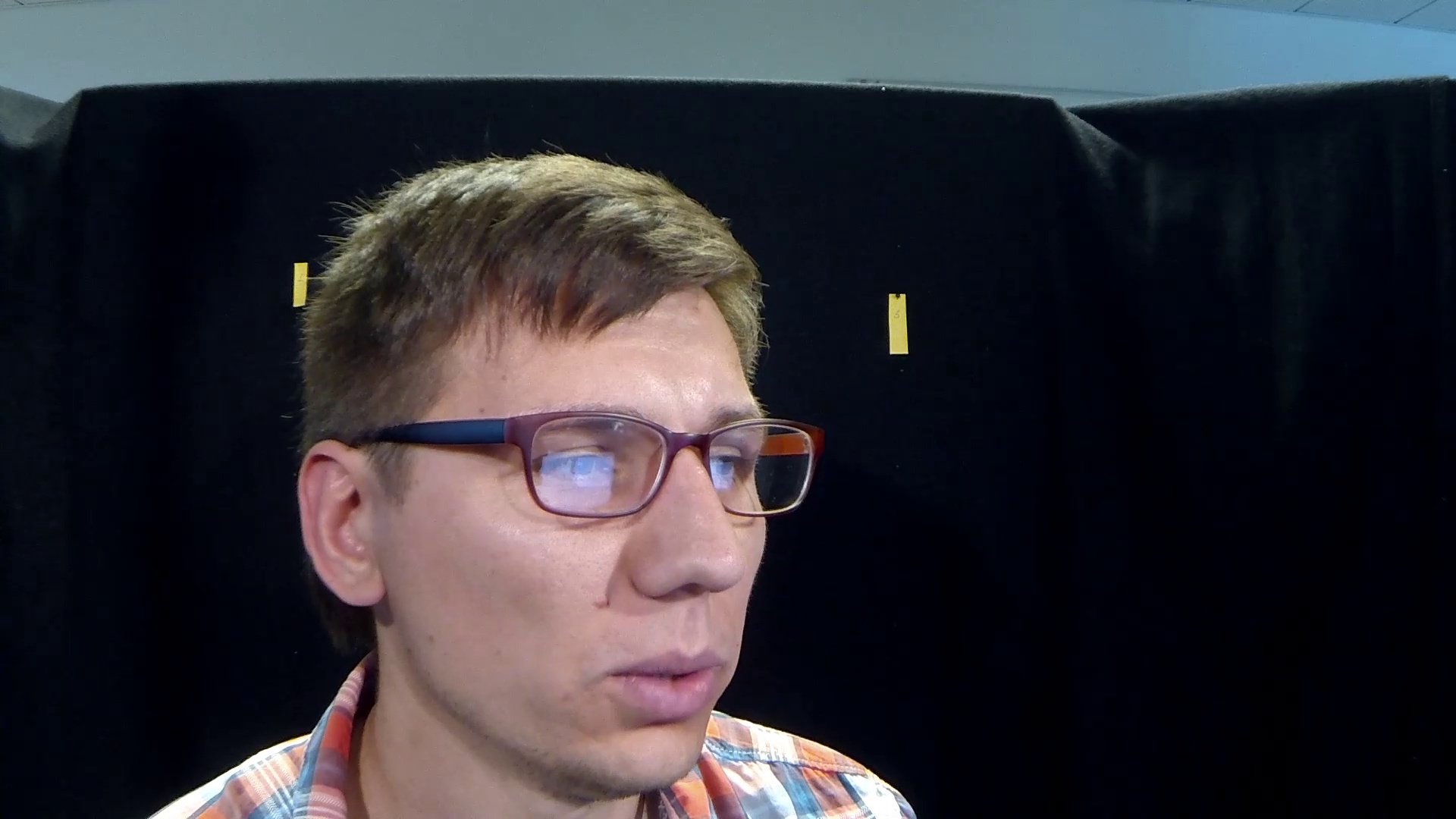}&
\includegraphics[trim = 9cm 1cm 27cm 6cm,clip,keepaspectratio=true,width=0.22\linewidth]{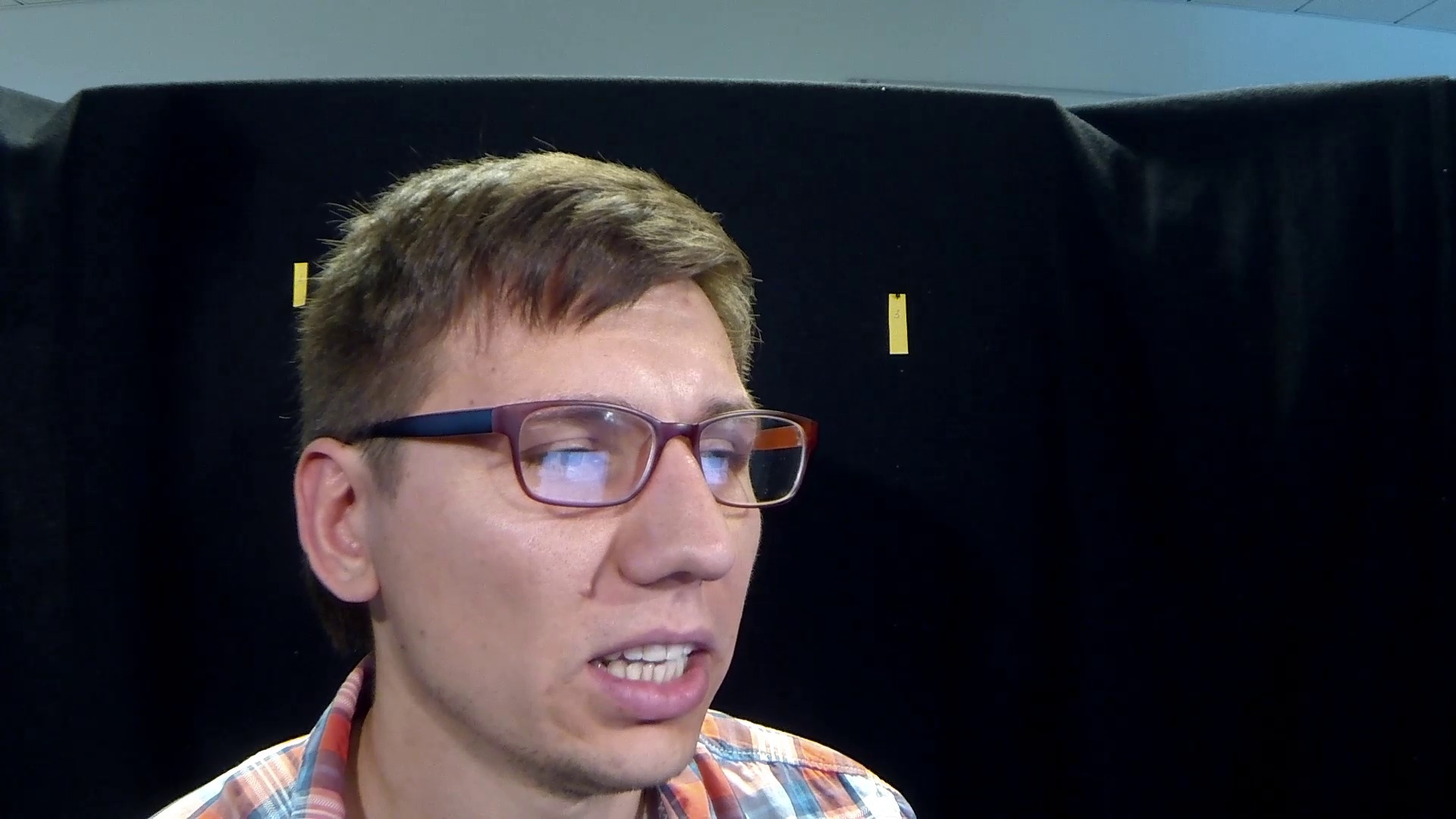}\\
%5(b)&
\includegraphics[trim = 20cm 6cm 20cm 6cm,clip,keepaspectratio=true,width=0.22\linewidth]{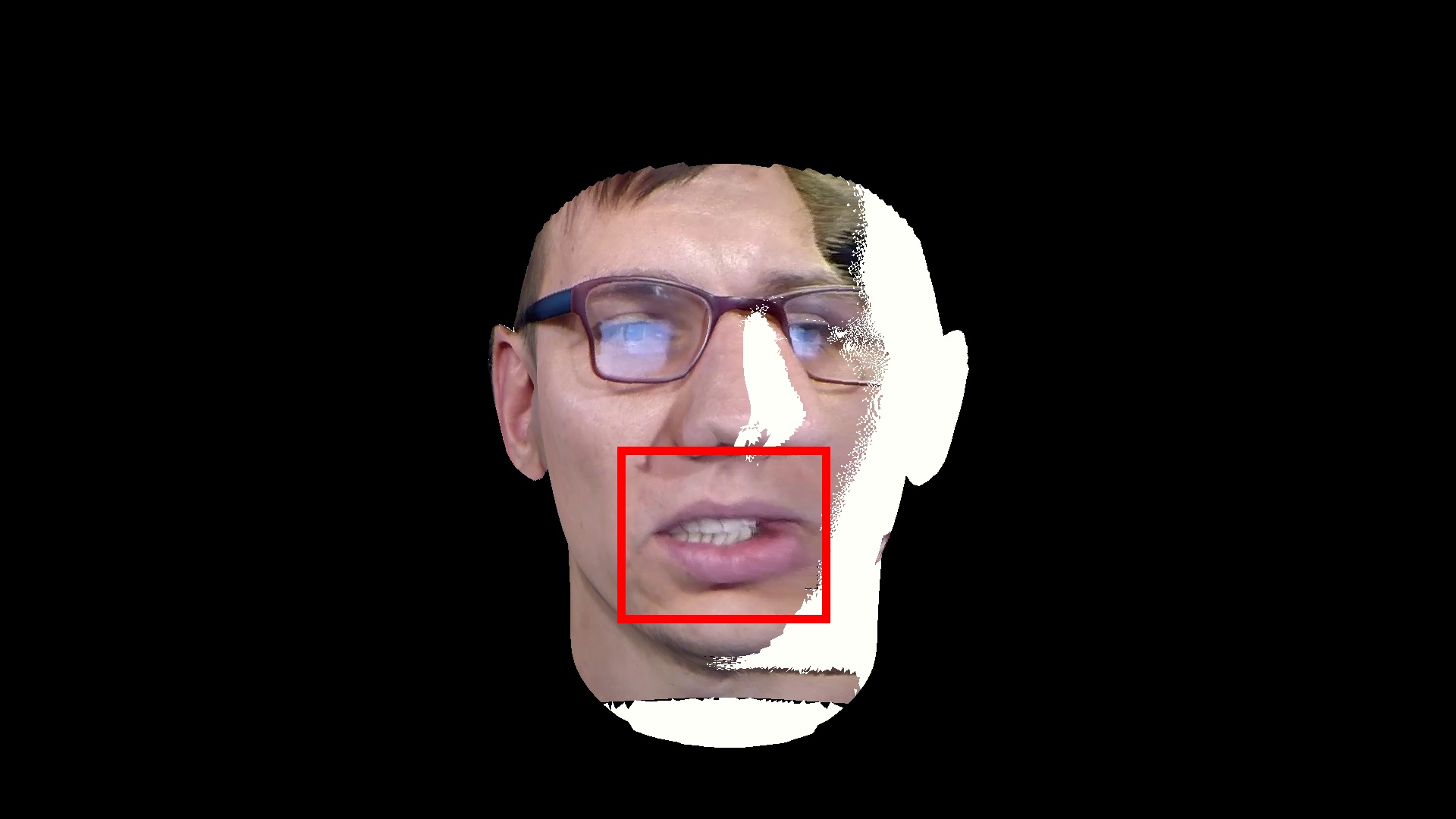}&
\includegraphics[trim = 20cm 6cm 20cm 6cm,clip,keepaspectratio=true,width=0.22\linewidth]{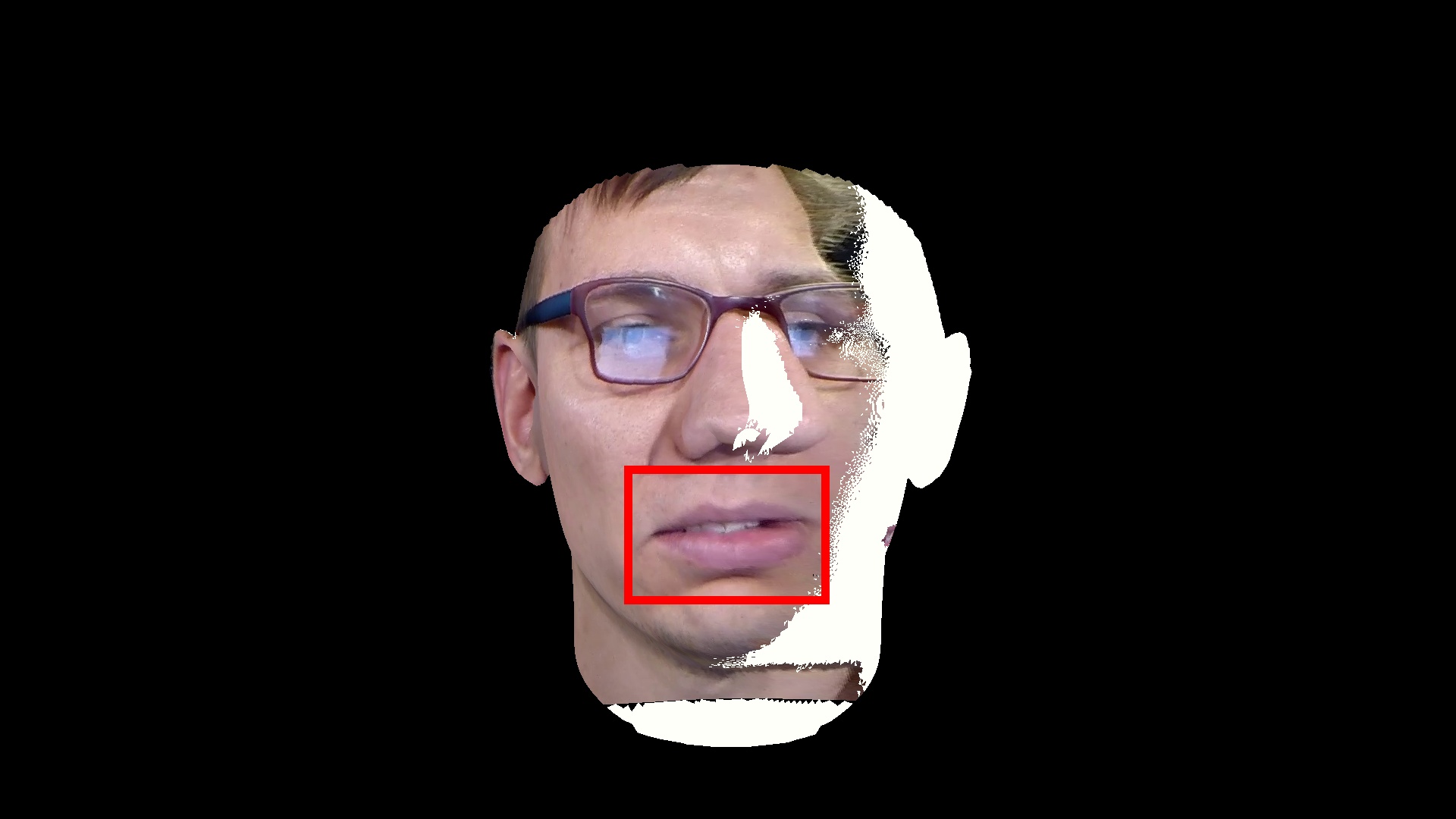}&
\includegraphics[trim = 20cm 6cm 20cm 6cm,clip,keepaspectratio=true,width=0.22\linewidth]{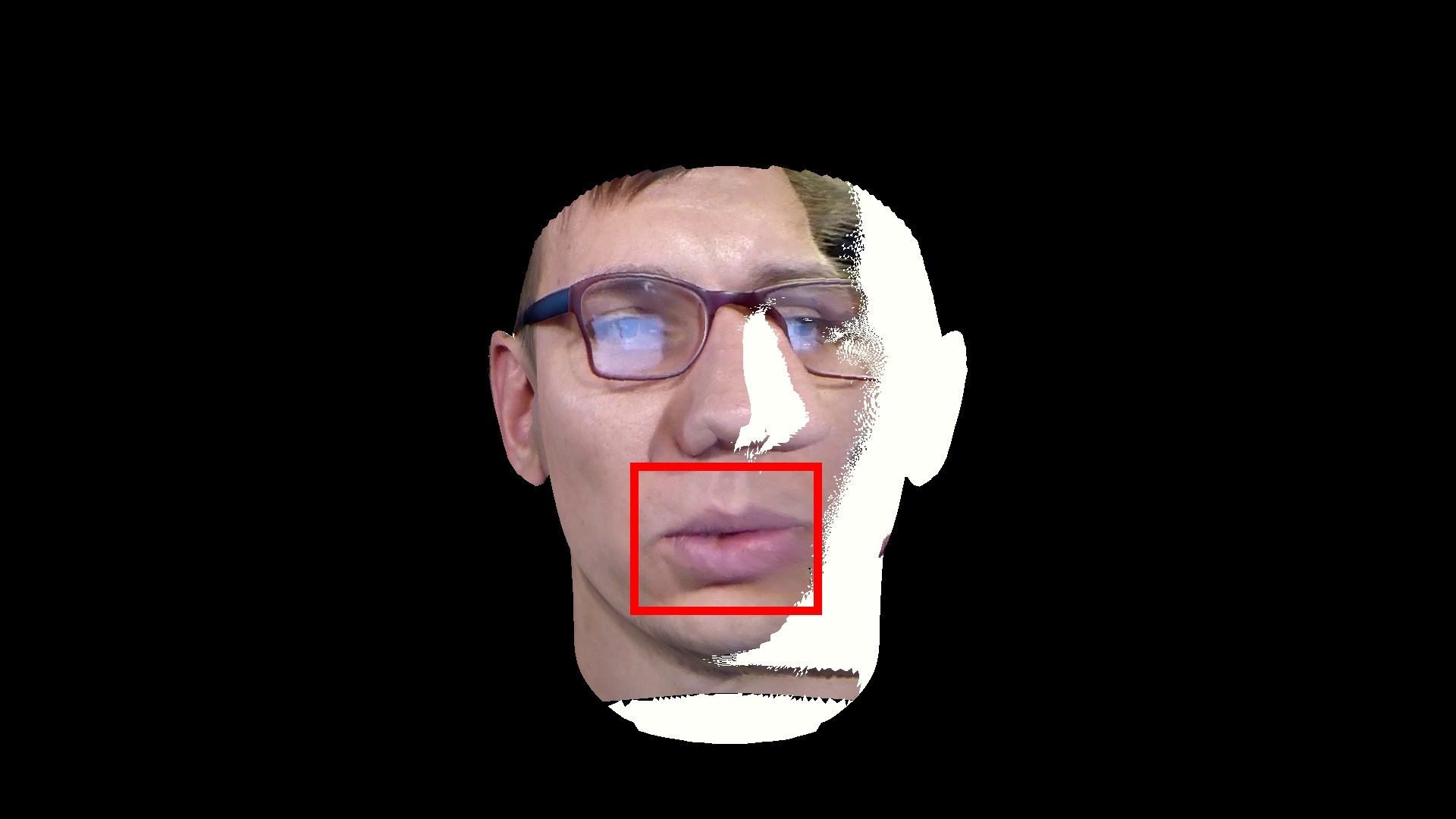}&
\includegraphics[trim = 20cm 6cm 20cm 6cm,clip,keepaspectratio=true,width=0.22\linewidth]{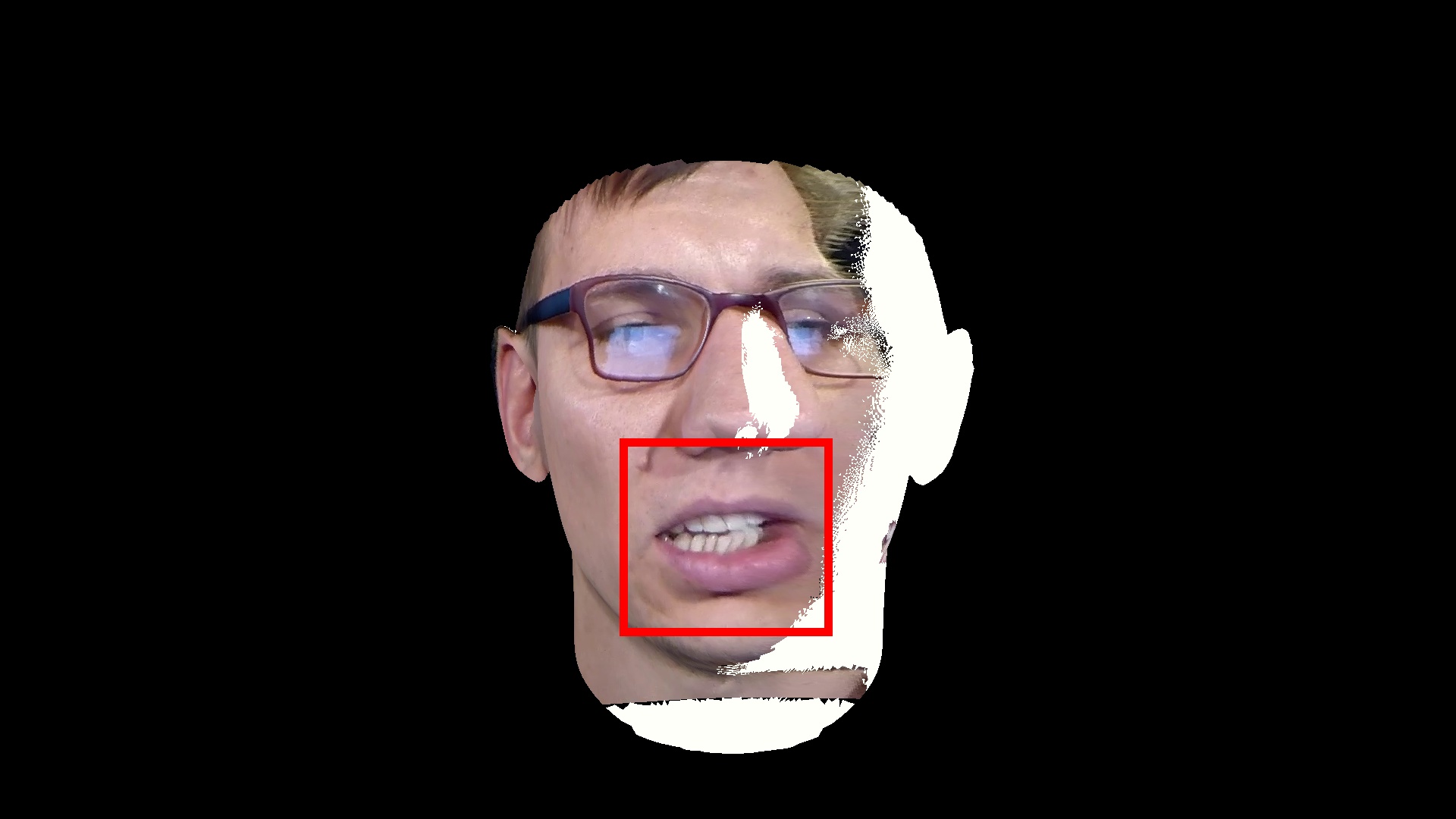}\\
0.707&0.778&0.818&0.663\\
%(c) &
\includegraphics[trim = 19cm 2cm 19cm 6cm,clip,keepaspectratio=true,width=0.22\linewidth]{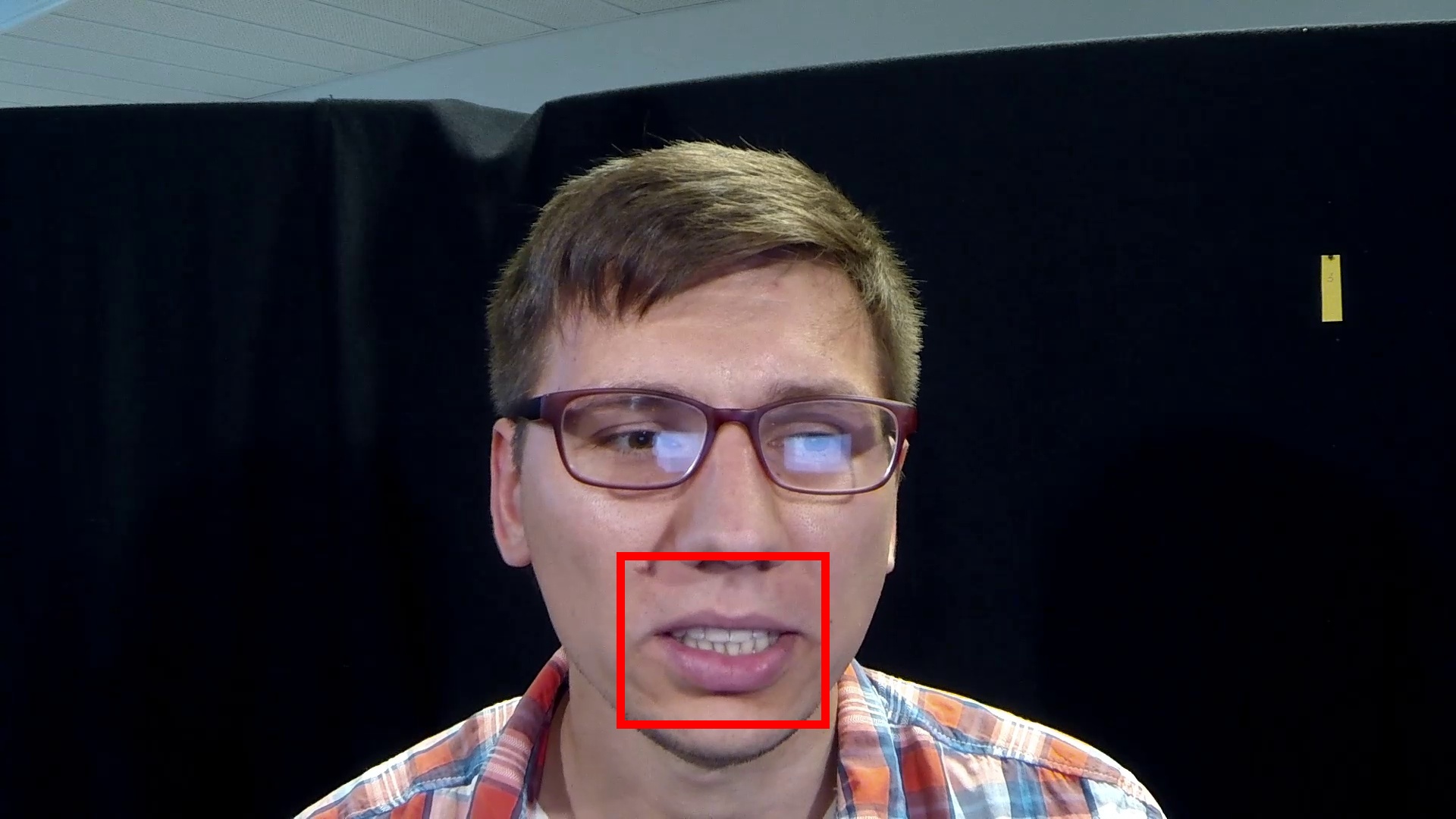}&
\includegraphics[trim = 19cm 2cm 19cm 6cm,clip,keepaspectratio=true,width=0.22\linewidth]{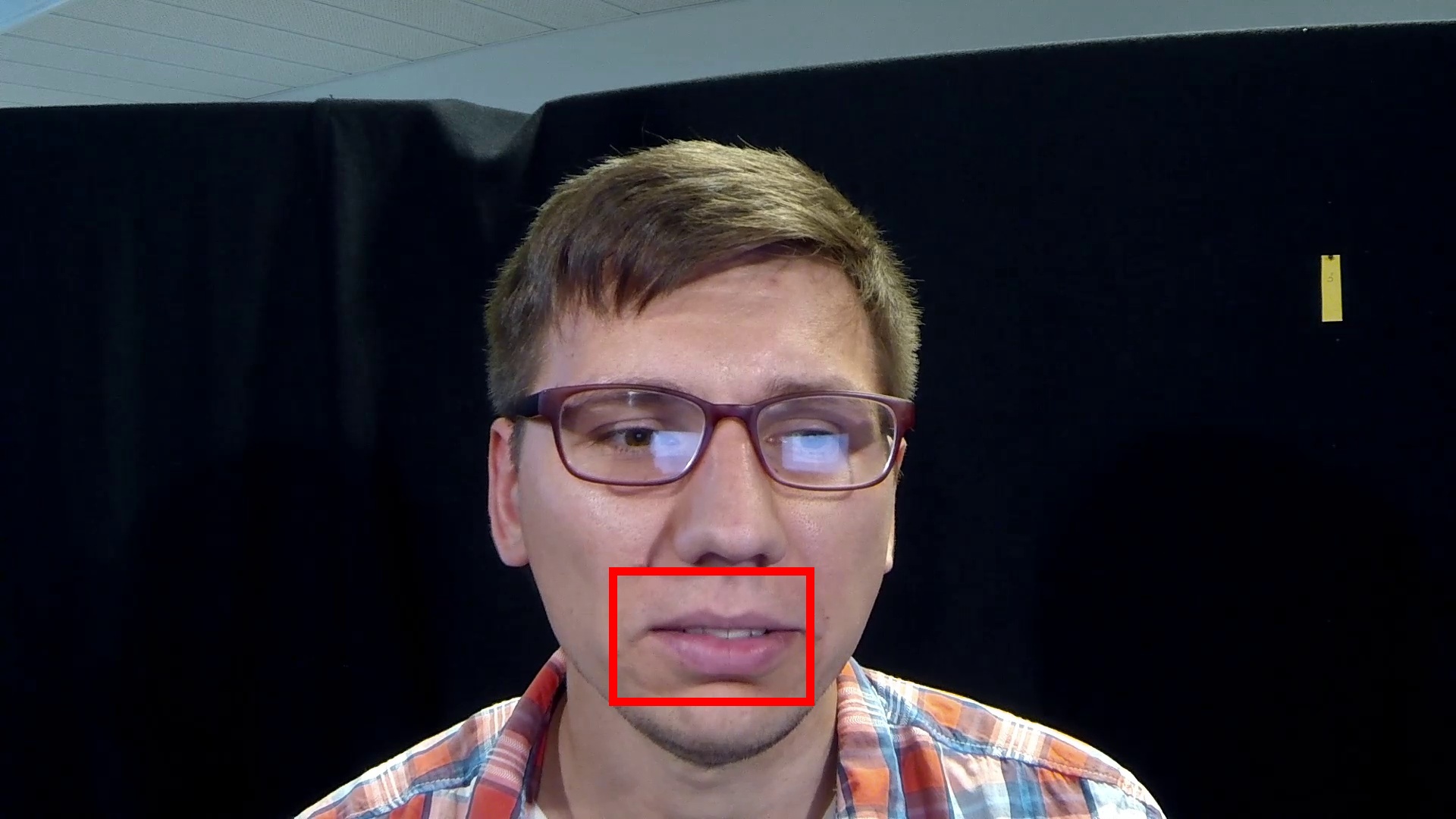}&
\includegraphics[trim = 19cm 2cm 19cm 6cm,clip,keepaspectratio=true,width=0.22\linewidth]{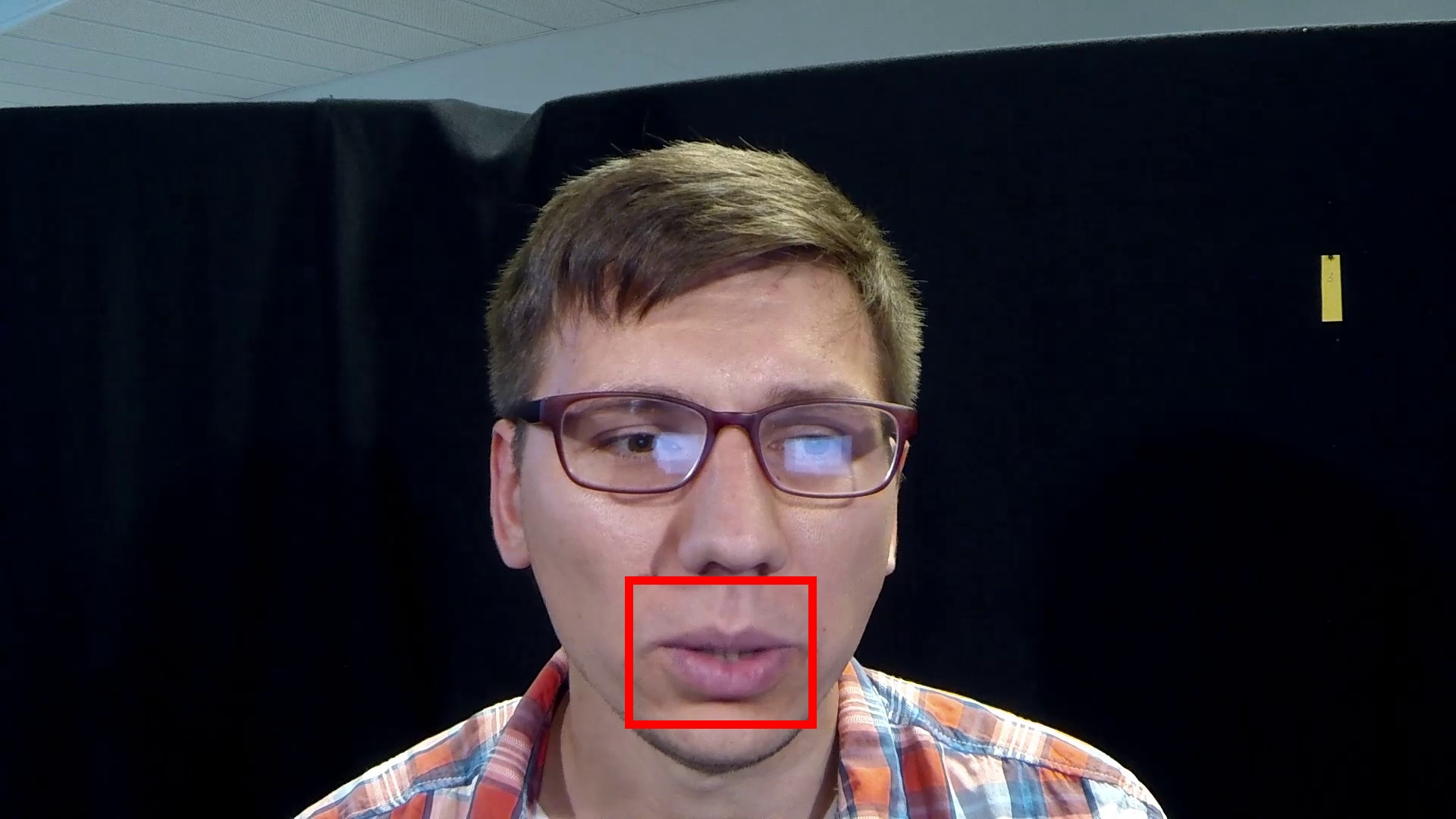}&
\includegraphics[trim = 19cm 2cm 19cm 6cm,clip,keepaspectratio=true,width=0.22\linewidth]{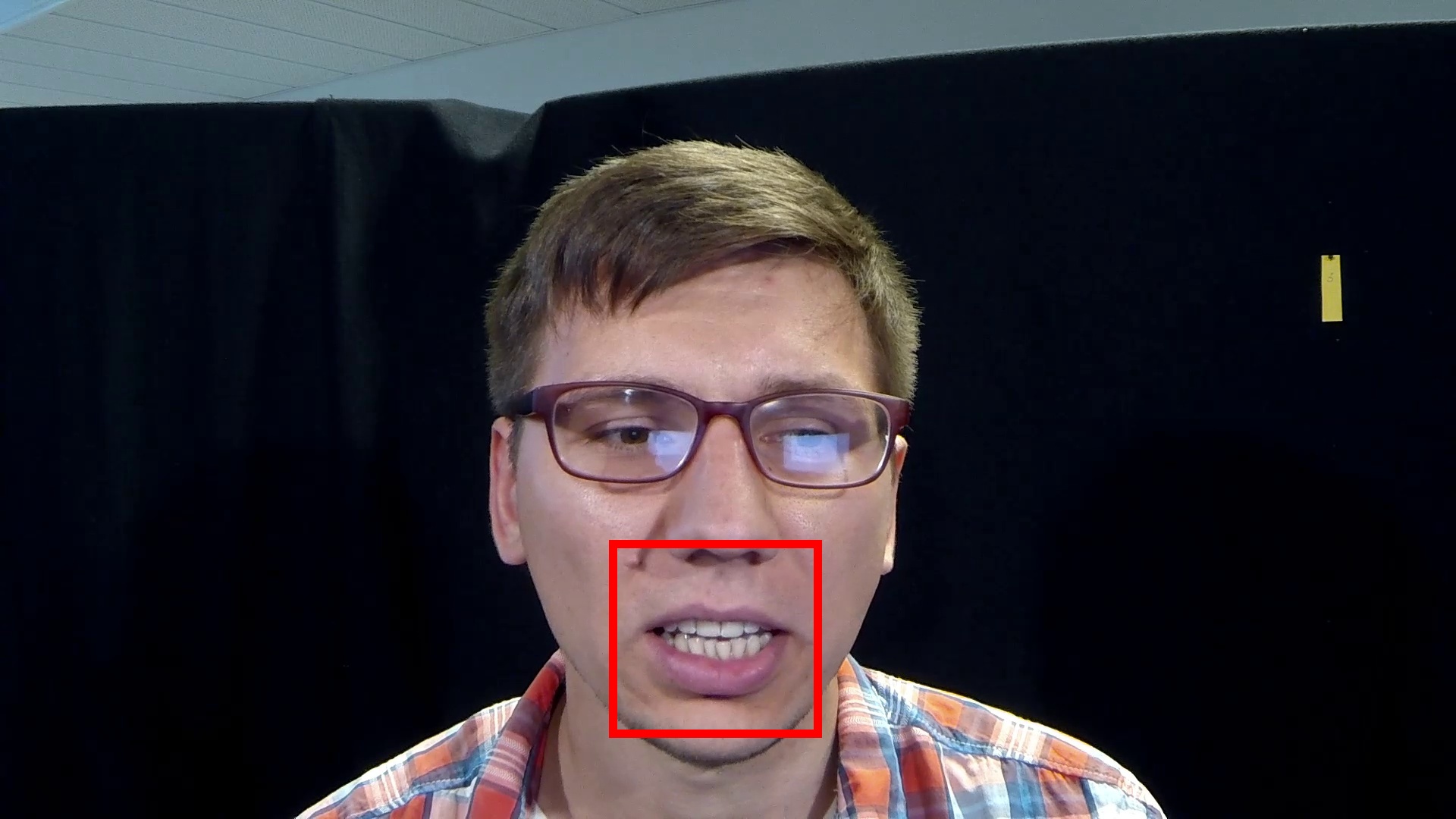}
%(d)&
\end{tabular}
    \caption{Same as Figure~\ref{fig:Oulu2} for participant \#21. The estimated horizontal head orientation (yaw angle) is of $40^\circ$ in this case. Notice that in spite of the large yaw angle, the mouth area correctly frontalized with no self occluded regions.}
    \label{fig:Oulu21}
\end{figure*}

\begin{figure*}[p!]
    \centering
\begin{tabular}{cccc}
\includegraphics[width=0.22\linewidth]{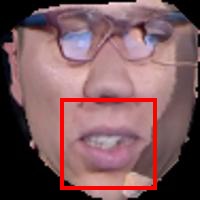}&
\includegraphics[width=0.22\linewidth]{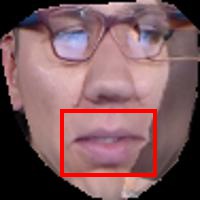}&
\includegraphics[width=0.22\linewidth]{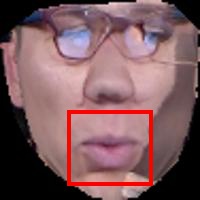}&
\includegraphics[width=0.22\linewidth]{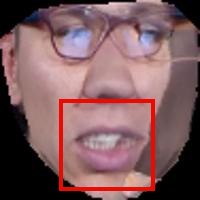}\\
0.572&0.732&0.795&0.586\\
\includegraphics[width=0.22\linewidth]{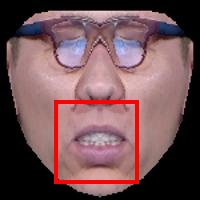}&
\includegraphics[width=0.22\linewidth]{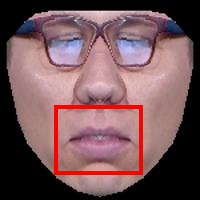}&
\includegraphics[width=0.22\linewidth]{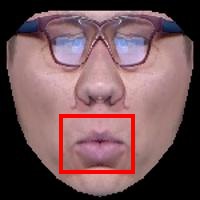}&
\includegraphics[width=0.22\linewidth]{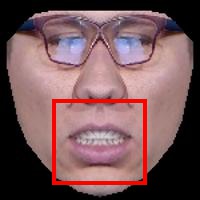}\\
0.668&0.723&0.833&0.743
\end{tabular}
    \caption{Same as Figure~\ref{fig:Oulu_Hassner_Banerjee_2} for participant \#21. Note that in this example symmetry post-processing introduces important facial distorsions. }
    \label{fig:Oulu_Hassner_Banerjee_21}
\end{figure*}

\begin{figure*}[t!]
    \centering
\begin{tabular}{cccc}
\includegraphics[trim = 15cm 5cm 15cm 0cm,clip,keepaspectratio=true,width=0.22\linewidth]{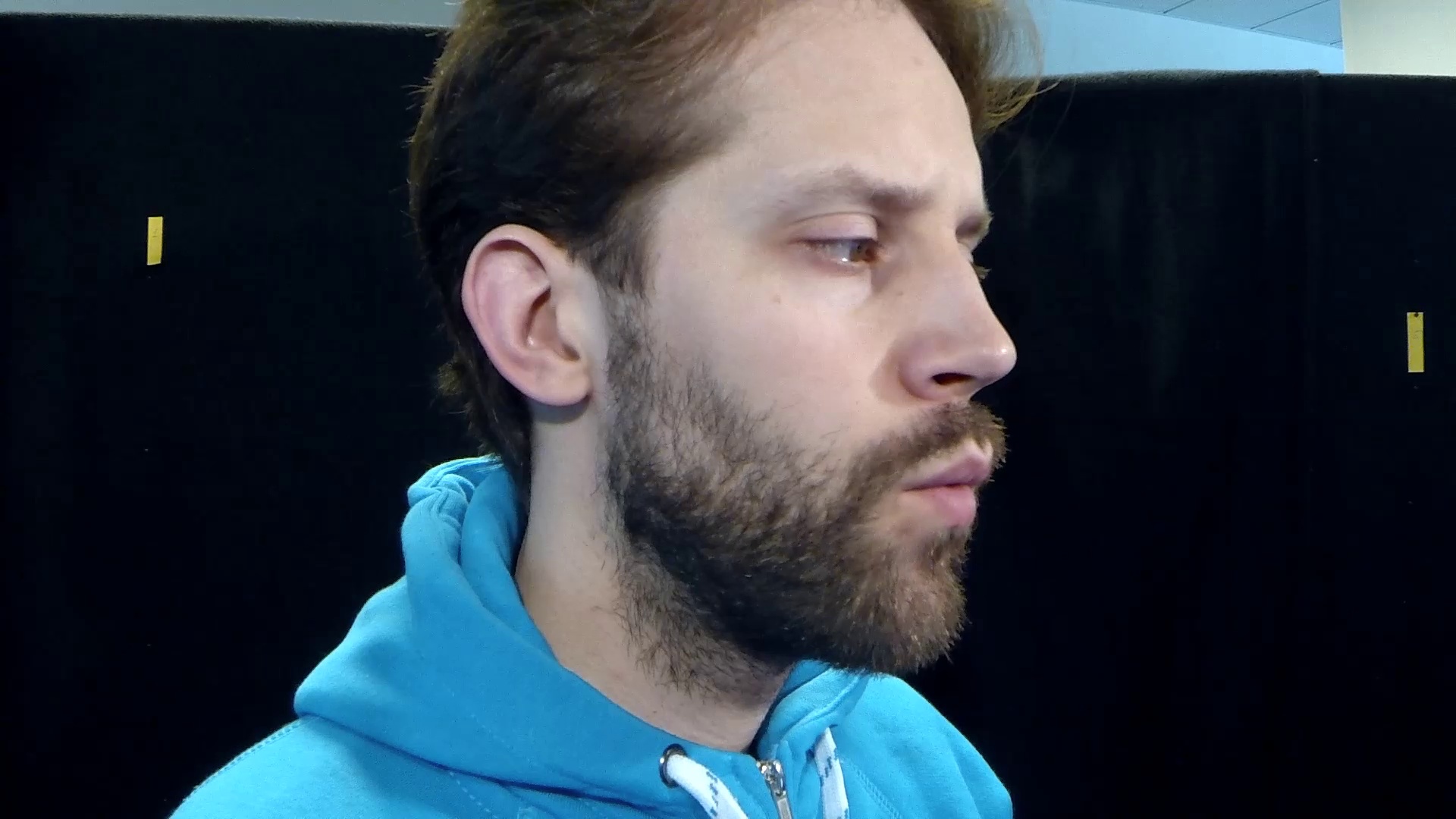}&
\includegraphics[trim = 15cm 2cm 15cm 3cm,clip,keepaspectratio=true,width=0.22\linewidth]{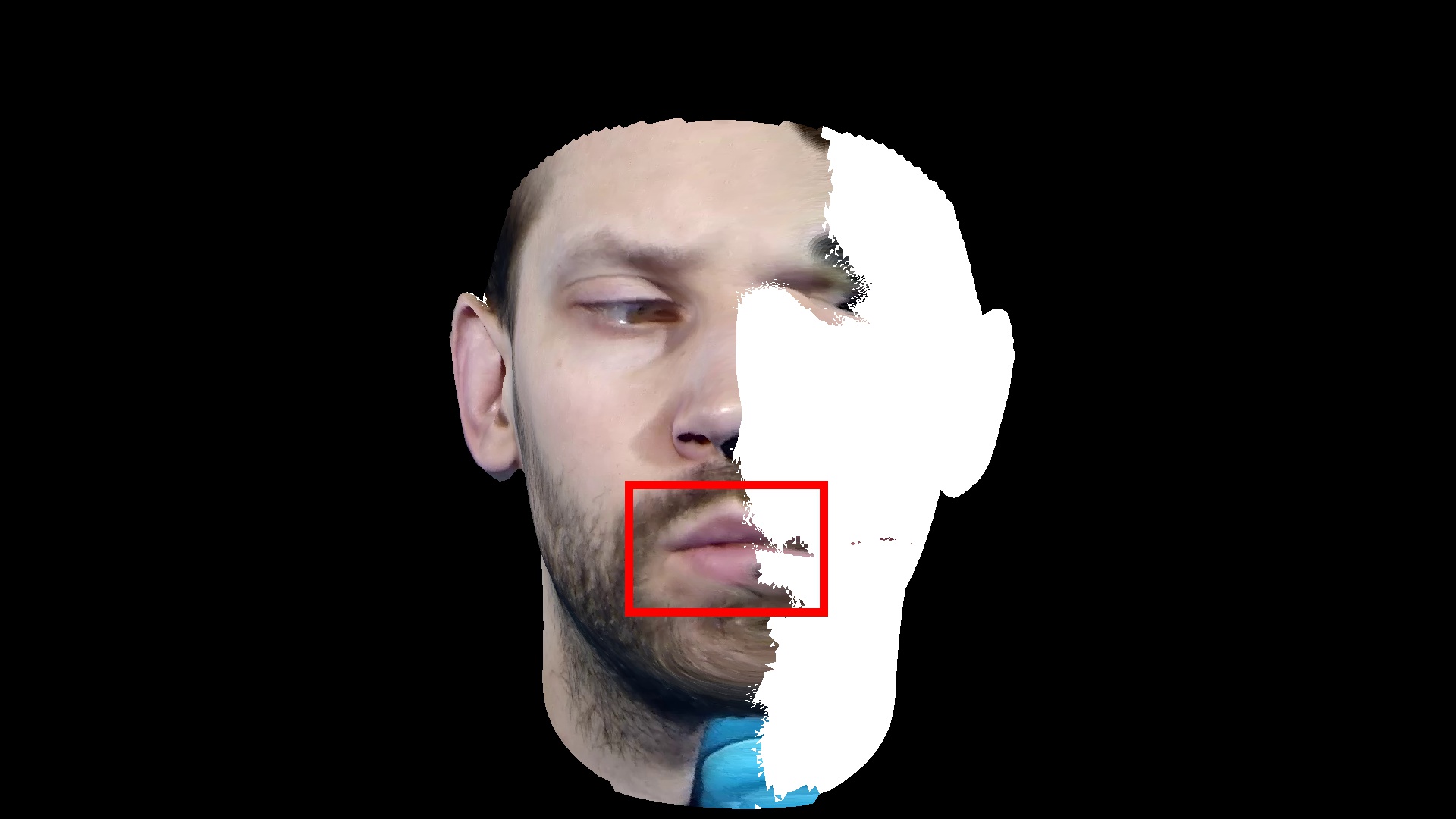}&
\includegraphics[trim = 15cm 2cm 15cm 3cm,clip,keepaspectratio=true,width=0.22\linewidth]{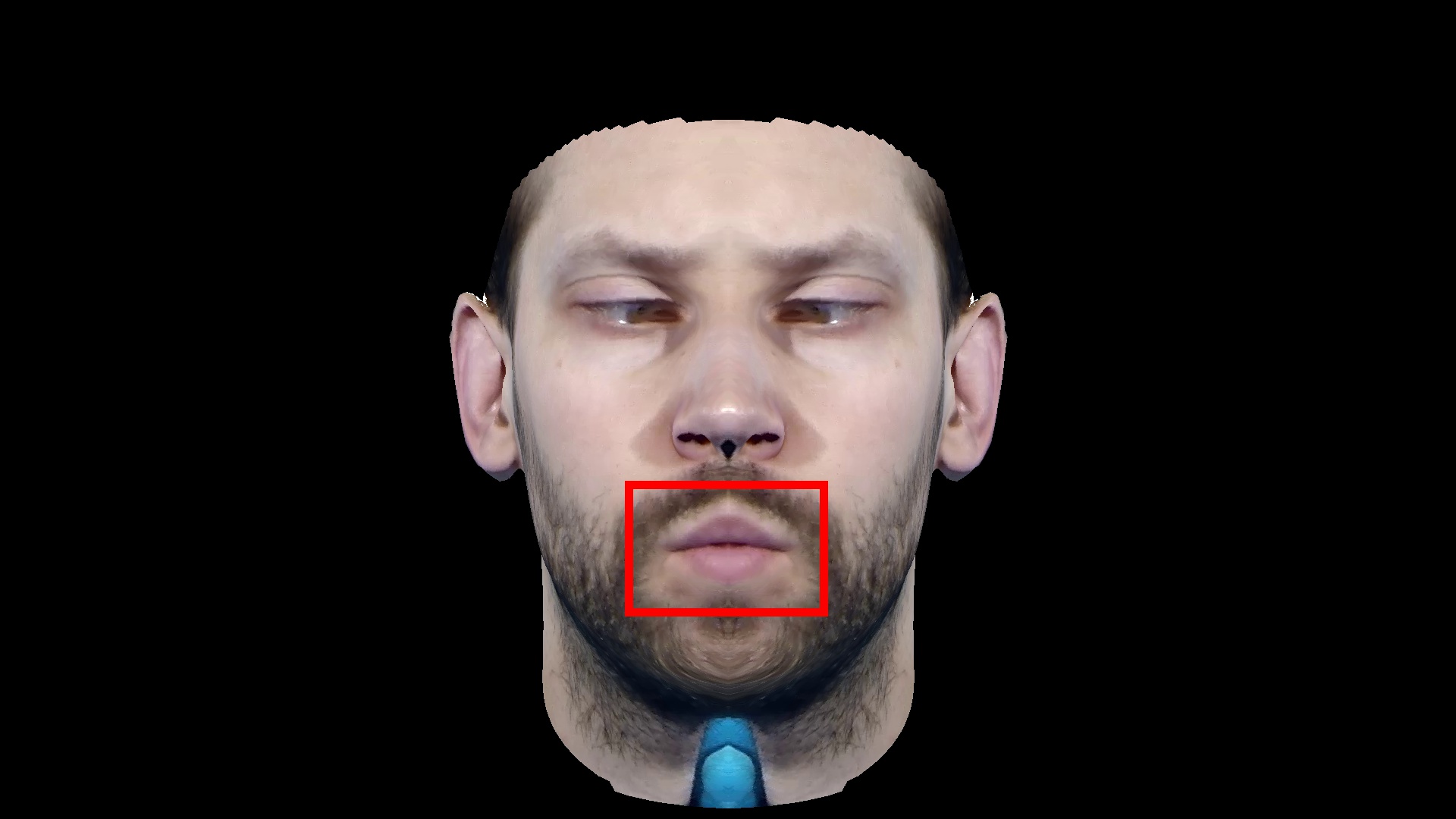}&
\includegraphics[trim = 15cm 5cm 15cm 0cm,clip,keepaspectratio=true,width=0.22\linewidth]{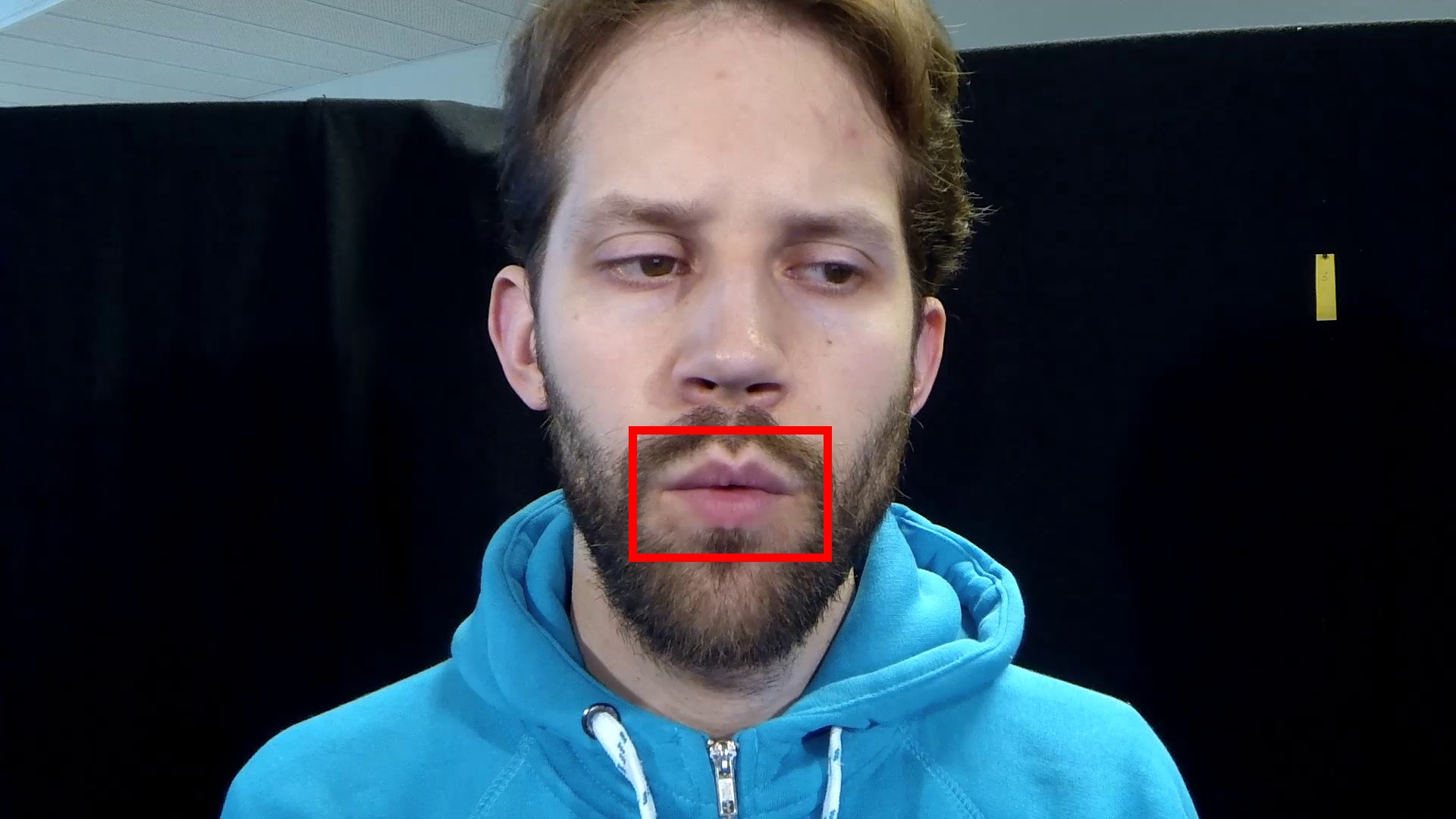}\\
(a) Input image & (b) 0.403 & (c) 0.739 (hard symmetry) & (d) ground truth 

\end{tabular}
    \caption{An example of face frontalization result obtained with the $45^\circ$ camera. (a) The estimated yaw angle is $57^\circ$ in this case. (b) Because the lip area is half occluded, the ZNCC coefficient is below 0.5. (c) Exploiting hard symmetry (the occluded part is filled in by flipping the visible part around the face's mirror-symmetric axis). (d) Ground-truth provided by the $0^\circ$ camera. }
    \label{fig:45degree-camera}
\end{figure*}

In practice, we evaluated the performance of the proposed method and we compared it with two state-of-the-art methods for which the code is publicly available, namely \cite{hassner2015effective,banerjee2018frontalize}. We applied frontalization to images extracted from the videos recorded with the $30^{\circ}$ camera and compared the results with the ``ground-truth", namely the corresponding images extracted from the videos recorded with the $0^{\circ}$ camera. Notice that videos recorded with higher viewing angles, i.e. $45^{\circ}$,  $60^{\circ}$ and $90^{\circ}$, can be hardly exploited by a frontalization algorithm because half of the face is occluded, e.g. Figure~\ref{fig:45degree-camera}. For each frontalized image $I_f$ we extract the mouth region $R_f$ and we search in the ground-truth image $I_t$ for the best-matching region  $R_t$. This provides a ZNCC coefficient \eqref{eq:zncc} for each test image. Notice that \eqref{eq:zncc} only cares about the horizontal and vertical shifts in the image plane and assumes that the frontalized face and the corresponding ground-truth frontal face share the same scale. In practice, different frontalization algorithms output faces at different scales. For this reason and for the sake of fairness, prior to applying  \eqref{eq:zncc}, we extract facial landmarks from both the frontalized and ground-truth faces and we use a subset of this set of landmarks to estimate the scale factor between the two faces. 

We randomly selected 30 video pairs, recorded with the $30^\circ$ and $0^\circ$ cameras, of 15 participants from the OulouVS2 dataset.
Each video contains approximatively 160 images, therefore we used a total of 4800 images in our benchmark. The mean ZNCC coefficients obtained with two state-of-the-art methods and with the proposed method are displayed in Table~\ref{table:zncc}. Both \cite{hassner2015effective} and \cite{banerjee2018frontalize} exploit facial symmetry to fill in the occluded areas, which improves the ZNCC metrics. Indeed, facial symmetry can be used, either by replacing occluded pixels with the mirror-symmetric ones (soft symmetry), or by flipping the visible half of the face (hard symmetry).

\begin{figure*}[t!]
    \centering
\begin{tabular}{cccc}
\includegraphics[width=0.20\linewidth]{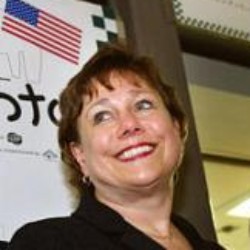}&
\includegraphics[trim = 0.4cm 0.4cm 0.4cm 0.4cm,clip,keepaspectratio=true,width=0.20\linewidth]{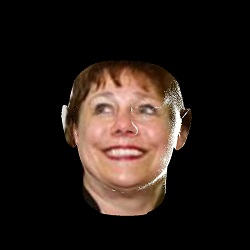}&
\includegraphics[width=0.20\linewidth]{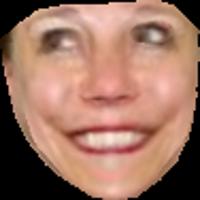}&
\includegraphics[width=0.20\linewidth]{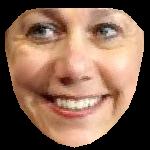}\\
\includegraphics[width=0.20\linewidth]{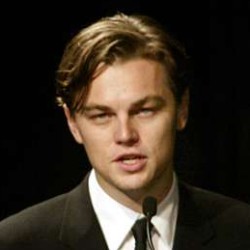}&
\includegraphics[trim = 0.4cm 0.4cm 0.4cm 0.4cm,clip,keepaspectratio=true,width=0.20\linewidth]{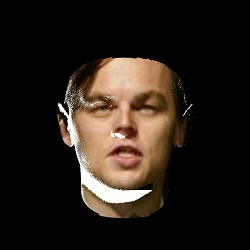}&
\includegraphics[width=0.20\linewidth]{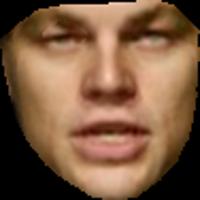}&
\includegraphics[width=0.20\linewidth]{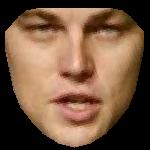}\\
\includegraphics[width=0.20\linewidth]{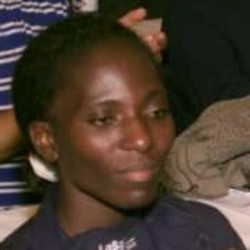}&
\includegraphics[trim = 0.4cm 0.4cm 0.4cm 0.4cm,clip,keepaspectratio=true,width=0.20\linewidth]{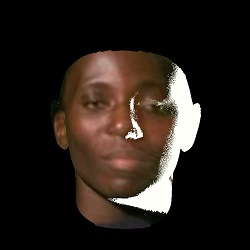}&
\includegraphics[width=0.20\linewidth]{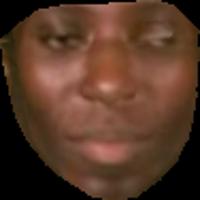}&
\includegraphics[width=0.20\linewidth]{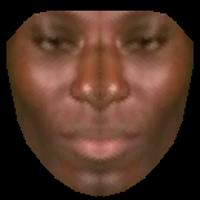}\\
\includegraphics[width=0.20\linewidth]{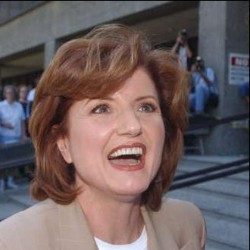}&
\includegraphics[trim = 0.4cm 0.4cm 0.4cm 0.4cm,clip,keepaspectratio=true,width=0.20\linewidth]{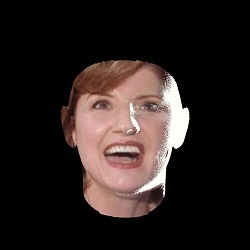}&
\includegraphics[width=0.20\linewidth]{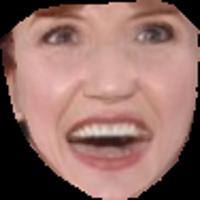}&
\includegraphics[width=0.20\linewidth]{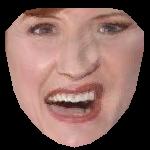} \\
(a) Input image & (b) Proposed & (c) Hassner et al \cite{hassner2015effective} & (d) Banerjee et al \cite{banerjee2018frontalize}
\end{tabular}
    \caption{Face frontalization examples from the Labeled Face in the Wild (LFW) dataset \cite{huang2008labeled}. Both \cite{hassner2015effective} and \cite{banerjee2018frontalize} make use of mirror symmetry, which introduces unrealistic facial deformations.}
    \label{fig:LFW_success}
\end{figure*}

We noticed that there were important discrepancies in method performance across participants. In order to better understand this phenomenon, we computed the mean ZNCC coefficient for nine participants and displayed these means and standard deviations as a function of the yaw angle (in degrees), i.e. the horizontal head orientation associated with the 3D head-pose estimated with the proposed method, please refer to Table~\ref{table:yawangles}. One may notice that there is a wide range of yaw angles, from $19^\circ$ and up to $40^\circ$, and that the performance gracefully decreases as the yaw angle increases. Nevertheless, the proposed method yields the best scores for participants \#19 and \#12 and the second best score for participant \#21, and this is without using mirror-symmetric information.
 
Examples of face frontalization obtained with our method and with the methods of \cite{hassner2015effective} and \cite{banerjee2018frontalize} are shown on Figure~\ref{fig:Oulu2}, Figure~\ref{fig:Oulu2}, Figure~\ref{fig:Oulu_Hassner_Banerjee_2} and Figure~\ref{fig:Oulu_Hassner_Banerjee_21}. As already mentioned, both \cite{hassner2015effective} and \cite{banerjee2018frontalize} enforce facial symmetry as a post-processing frontalization step to compensate for the gaps caused by self occlusions. We also show an example of applying frontalization to the $45^\circ$ camera, Figure~\ref{fig:45degree-camera}. In this example the estimated yaw angle is $57^\circ$ and only half of the lip area is visible. Consequently, the ZNCC coefficient is below 0.5 in this case. Notice that when the hard-symmetry strategy of \cite{hassner2015effective} is applied, the ZNCC coefficient increases considerably.

\begin{figure*}[h!]
    \centering
\begin{tabular}{cccc}
\includegraphics[width=0.20\linewidth]{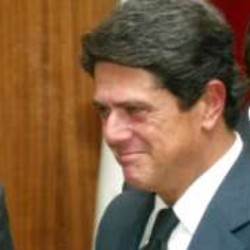}&
\includegraphics[trim = 0.4cm 0.4cm 0.4cm 0.4cm,clip,keepaspectratio=true,width=0.20\linewidth]{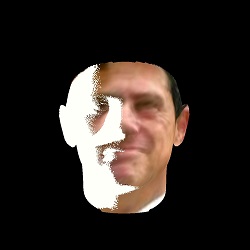}&
\includegraphics[trim = 0.4cm 0.4cm 0.4cm 0.4cm,clip,keepaspectratio=true,width=0.20\linewidth]{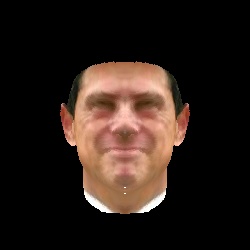}&
\includegraphics[width=0.20\linewidth]{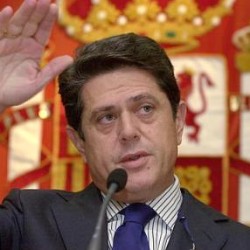}\\
\includegraphics[width=0.20\linewidth]{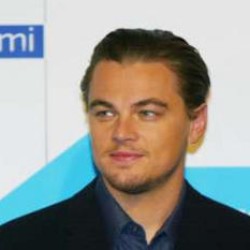}&
\includegraphics[trim = 2cm 2cm 2cm 2cm,clip,keepaspectratio=true,width=0.20\linewidth]{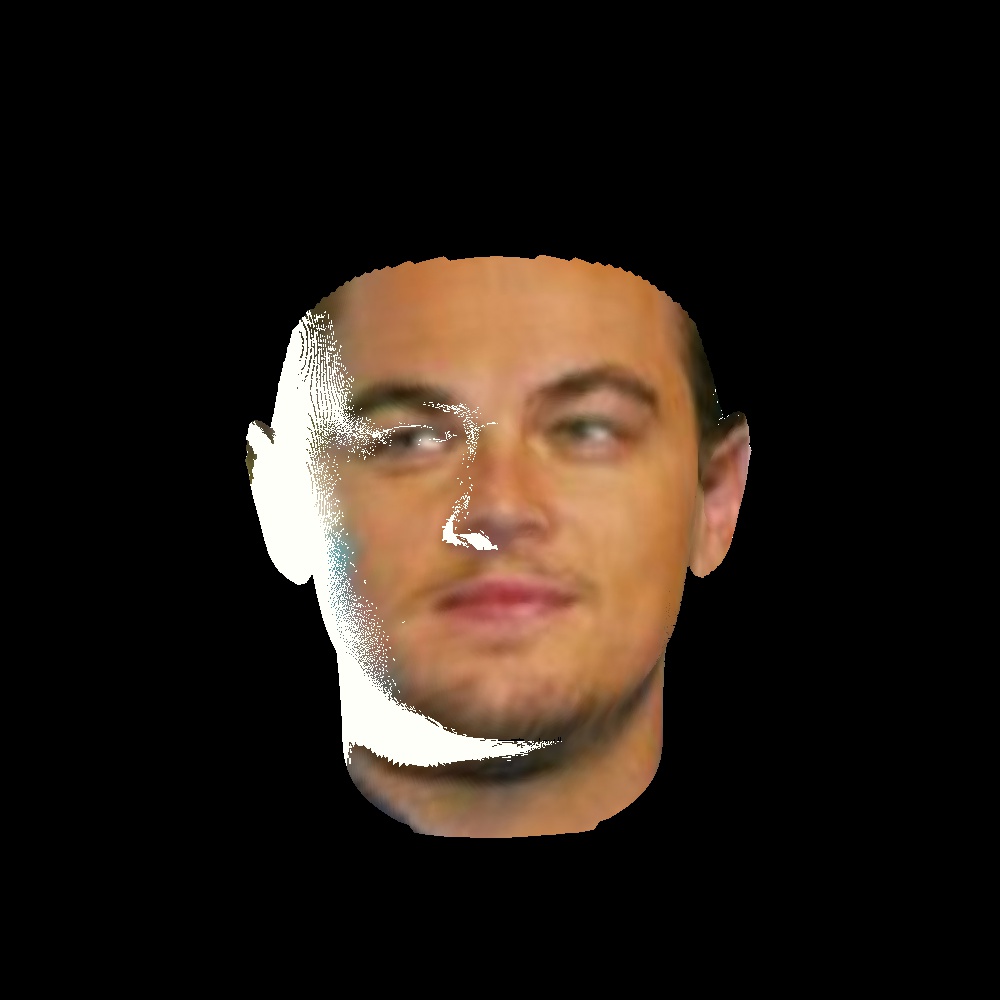}&
\includegraphics[trim = 2cm 2cm 2cm 2cm,clip,keepaspectratio=true,width=0.20\linewidth]{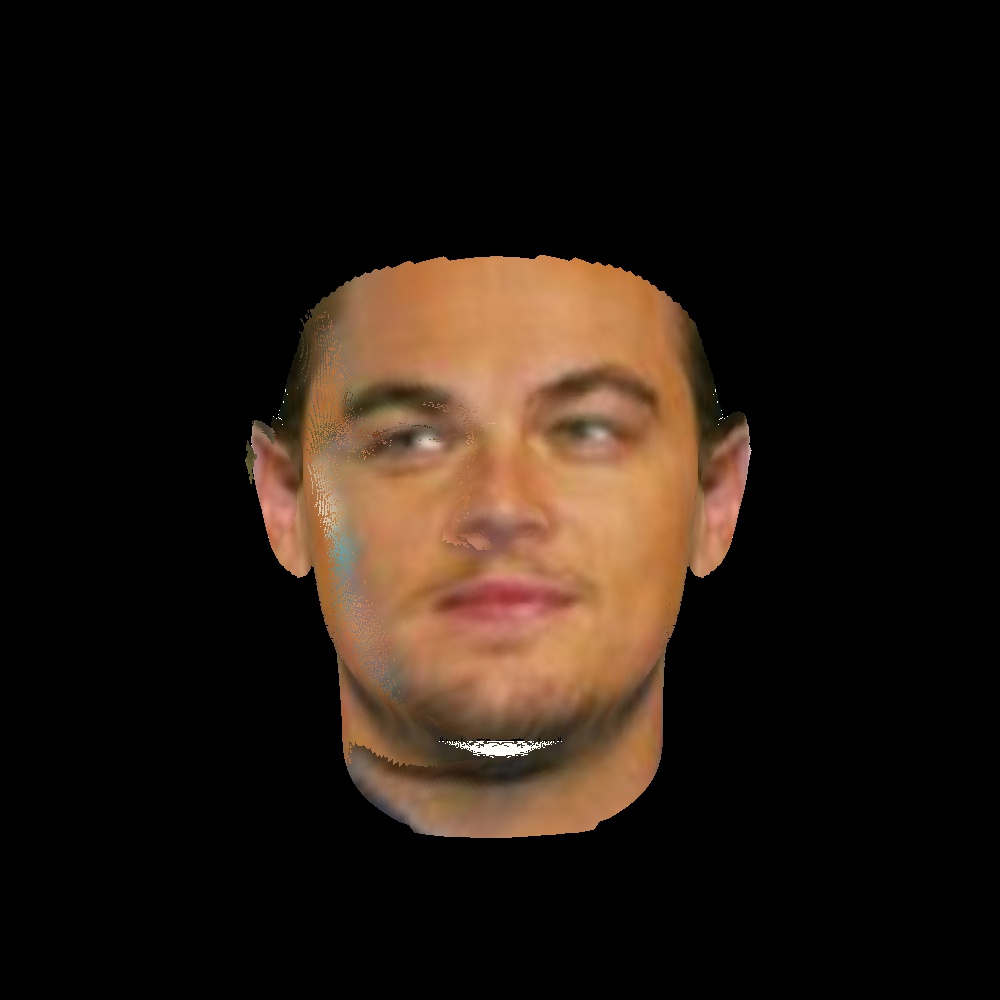}&
\includegraphics[width=0.20\linewidth]{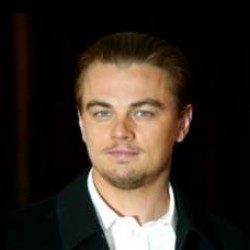}\\
\includegraphics[width=0.20\linewidth]{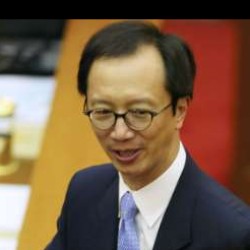}&
\includegraphics[trim = 2cm 2cm 2cm 2cm,clip,keepaspectratio=true,width=0.20\linewidth]{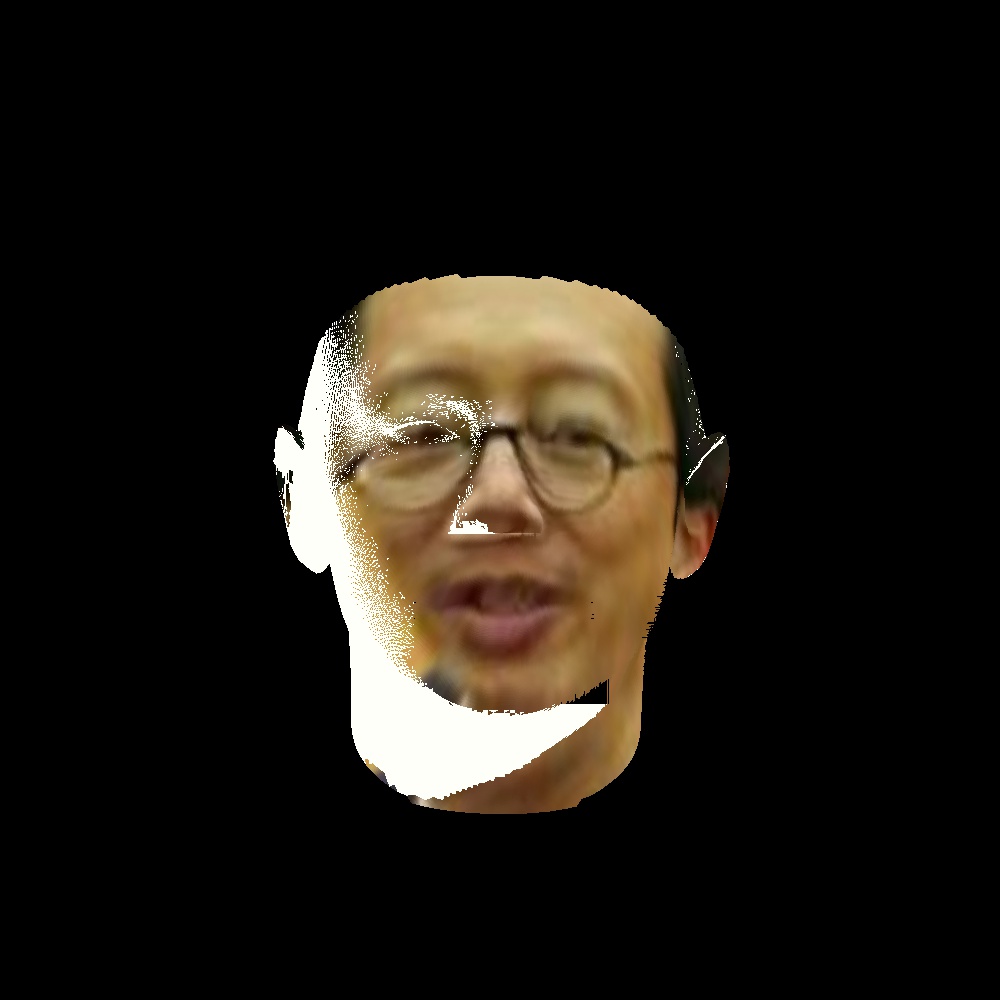}&
\includegraphics[trim = 2cm 2cm 2cm 2cm,clip,keepaspectratio=true,width=0.20\linewidth]{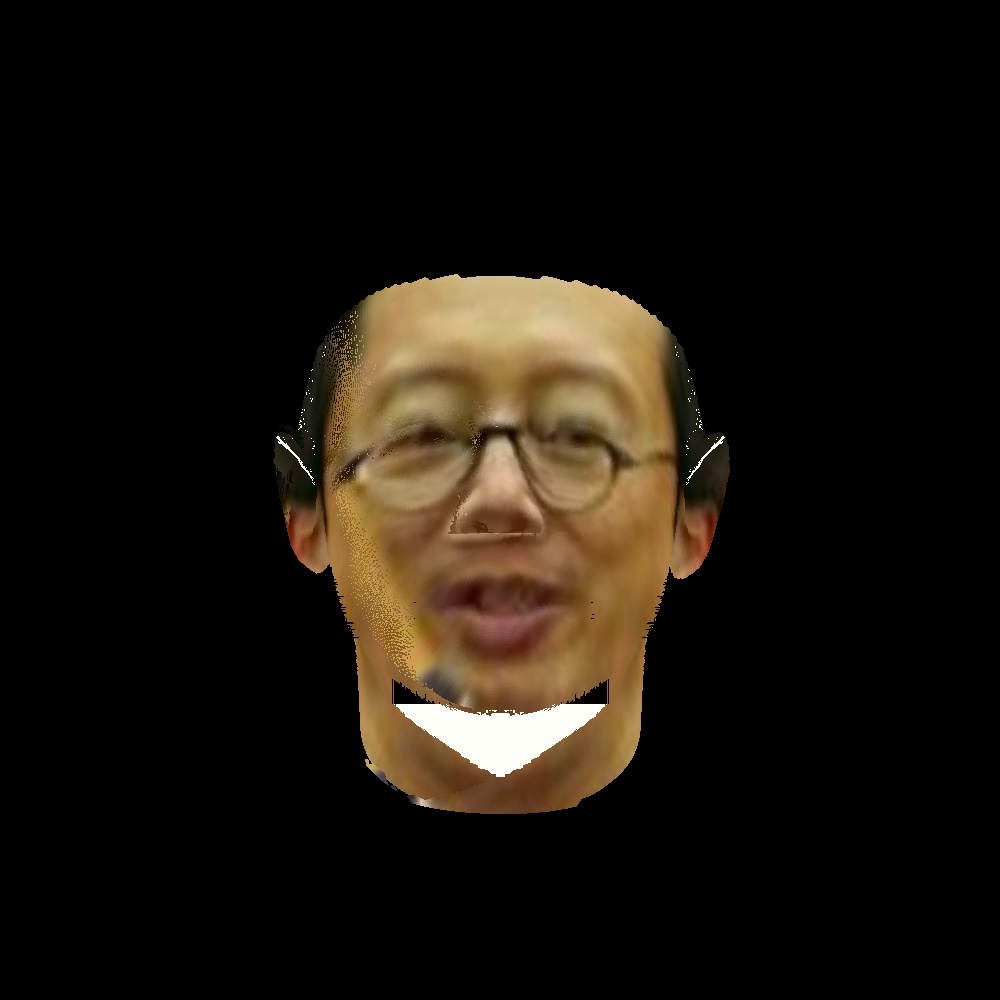}&
\includegraphics[width=0.20\linewidth]{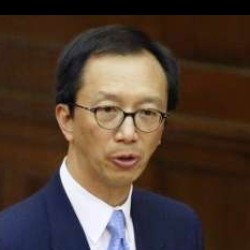}\\
(a) Input image & (b) Proposed & (c) Proposed + soft symmetry & (d) Frontal image
\end{tabular}
    \caption{Additional examples from LFW  showing the visual impact of using soft symmetry in conjunction with the proposed method. The frontal image (d) is shown for qualitative visual comparison and it is not used as ground truth.}
    \label{fig:LFW_sym}
\end{figure*}

Finally, Figure~\ref{fig:LFW_success} and Figure~\ref{fig:LFW_sym} show a few examples of frontalization results obtained with faces from the LFW dataset \cite{huang2008labeled}. It is worthwhile to notice that our method yields frontal images of a better quality than the methods of
\cite{hassner2015effective} and \cite{banerjee2018frontalize}, and that these two methods produce many artefacts and unrealistic facial deformations.

%\begin{figure*}[h!]
%    \centering
%\begin{tabular}{cccc}
%\includegraphics[width=0.22\linewidth]{eric.jpg}&
%\includegraphics[width=0.22\linewidth]{eric_our_visible.jpg}&
%\includegraphics[width=0.22\linewidth]{eric_Hassner.jpg}&
%\includegraphics[width=0.22\linewidth]{eric_Banerjee.jpg}\\
%
%\includegraphics[width=0.20\linewidth]{ann_bad.jpg}&
%\includegraphics[width=0.20\linewidth]{ann_bad_our_visible.jpg}&
%\includegraphics[width=0.20\linewidth]{ann_bad_Hassner.jpg}&
%\includegraphics[width=0.20\linewidth]{ann_bad_Banerjee.jpg}\\
%
%\includegraphics[width=0.20\linewidth]{federico.jpg}&
%\includegraphics[width=0.20\linewidth]{federico_our_visible.jpg}&
%\includegraphics[width=0.20\linewidth]{federico_Hassner.jpg}&
%\includegraphics[width=0.20\linewidth]{federico_Banerjee.jpg}\\
%
%\includegraphics[width=0.20\linewidth]{fred.jpg}&
%\includegraphics[width=0.20\linewidth]{fred_our_visible.jpg}&
%\includegraphics[width=0.20\linewidth]{fred_Hassner.jpg}&
%\includegraphics[width=0.20\linewidth]{fred_Banerjee.jpg}
%\end{tabular}
%    \caption{Same as \ref{fig:LFW_success} but with failure cases. First row: illumination difference. Second row: Distortion of lip region because of the head rotation. Third row: extreme pose. Fourth row: occlusion. } 
%    \label{fig:LFW_failure}
%\end{figure*}

\section{Conclusions}
\label{sec:conclusions}
In this paper we introduced a robust face frontalization (RFF) method that is based on the simultaneous estimation of the rigid transformation between two 3D point sets and the non-rigid deformation of a 3D face model. This is combined with pixel-to-pixel warping, between an input image of an arbitrarily-viewed face and a synthesized frontal view of the face. The proposed method yields state-of-the-art results, both quantitatively and qualitatively, when compared with two recently proposed methods. Up to now, the performance of face frontalization has been evaluated using face recognition benchmarks. We advocate that direct image-to-image comparison between the predicted output and the associated ground-truth yields a better assessment criterion that is not biased by another task, e.g. face recognition. 

It is worthwhile to notice that the performance of face recognition can be boosted by combining face frontalization with the use of facial symmetry. However, the latter is likely to introduce undesired artefacts and unrealistic facial deformations, which are quite damaging for other face analysis tasks, such as expression recognition or lip reading. 

In the future, we plan to extend the proposed method to image sequences in order to allow robust temporal analysis of faces. In particular, we are interested in combining face frontalization with audio-visual speech enhancement. As already outlined, face frontalization may well be viewed as a process of discriminating between rigid head movements and non-rigid facial deformations, which can be used to eliminate head motions that naturally accompany speech production, and hence leverage the performance of visual speech reading and of audio-visual speech enhancement.
\appendices
\section{Robust Estimation of the Rigid Alignment Between Two 3D Point Sets}
\label{app:robust-alignment}
In this appendix, we address the problem of robust alignement between two 3D point sets, $\mathcal{X}$ with coordinates $\Xvect_{1:J} = \{ \Xvect_j \}_{j=1}^J \subset \mathbb{R}^{3 \times J}$ and $\mathcal{Z}$ with coordinates $\Zvect_{1:J} = \{ \Zvect_j \}_{j=1}^J \subset \mathbb{R}^{3 \times J}$, respectively. As mentioned in Section~\ref{sec:frontalization-landmarks} the residuals \eqref{eq:rigid-residual} are samples of a random variable drawn from the generalized Student t-distribution \cite{forbes2014new}:
\begin{align}
\label{eq:generalized-student-app}
P(\evect; \Sigmamat, & \mu, \nu)   = \int_{0}^{\infty} \mathcal{N} ( \evect; 0, w \inverse \Sigmamat) \mathcal{G}(w, \mu, \nu) dw \nonumber \\
&=
\frac{\Gamma(\mu + \frac{3}{2})}{| \Sigmamat |^{\frac{1}{2}} \Gamma(\mu) (2\pi \nu)^{\frac{3}{2}}}
\left(
1+ \frac{\| \evect \|^2_{\mathcal{M}}} {2\nu}
\right)_{,}^{-\left(\mu + \frac{3}{2}\right)}
\end{align}
where $\mu$ and $\nu$ are the parameters of the \textit{prior gamma distribution} of the precision  variable $w$, and $\Gamma(\cdot)$ is the gamma function. Without loss of generality, we set $\nu=1$ \cite{forbes2014new}. Notice that using the conjugate prior property, the posterior distribution of $w$ is also a gamma distribution, namely the \textit{posterior gamma distribution}:
\begin{align}
\label{eq:weight-gamma-posterior}
P(w | \evect; \Sigmamat, \mu, \nu) &= \mathcal{N} ( \evect; 0, w \inverse \Sigmamat) \mathcal{G}(w, \mu, 1) \nonumber \\
& = \mathcal{G} (w; a, b),
\end{align}
with parameters:
\begin{align}
\label{eq:gamma-posterior-ab}
a  = \mu + \frac{3}{2}, \quad
%\label{eq:gamma-posterior-b}
b = 1 + \frac{1}{2} \| \evect \| ^2_{\Sigmamat}.
\end{align}
The posterior mean of the precision is:
\begin{equation}
\label{eq:weight-expectation}
\overline{w} = \mathrm{E}[w | \evect ] = \frac{a}{b}.
\end{equation}
Direct minimization of the likelihood function is not practical. An expectation-maximization formulation is therefore adopted, namely the minimization of the \textit{expected complete-data negative log-likelihood} \eqref{eq:expected-student}, $\mathrm{E}_W [ - \log P( \evect_{1:J}, w_{1:J} | \evect_{1:J} ; \thetavect)]$ and in this case the parameter vector is a concatenation of the alignment and pdf parameters, namely $\thetavect = \{\rho, \Rmat, \tvect, \Sigmamat, \mu\}$ since we set $\nu=1$. This yields the minimization of:
\begin{equation}
\label{eq:expectation-student}
\mathcal{Q}(\thetavect) =  \frac{1}{2} \sum_{j=1}^{J} \big( \overline{w}_j \| \Yvect_j - \rho \Rmat \Xvect_j - \tvect \|^2_{\Sigmamat} + \log |\Sigmamat| \big),
\end{equation}

By taking the derivatives of \eqref{eq:expectation-student} with respect to the parameters and equating to zero, we obtain the following analytical expressions for optimal values of the following parameters:
\begin{align}
\label{eq:opt-translation-robust}
\tvect^{\star} &= \overline{\Yvect} - \rho \Rmat \overline{\Xvect},\\
\label{eq:opt-scale-robust}
\rho^{\star} & = \left( \frac{\sum_{j=1}^{J} \overline{w}_j \Yvect_j^{\prime\top}  \Sigmamat\inverse \Yvect\pri_j}{\sum_{j=1}^{J} \overline{w}_j (\Rmat \Xvect_j\pri )\tp  \Sigmamat\inverse (\Rmat\Xvect\pri_j)} \right) ^{\frac{1}{2}}, \\
\label{eq:covariance-student}
\Sigmamat^{\star} &= \frac{1}{N} \sum_{j=1}^{J} \overline{w}_j (\Yvect\pri_j - \rho \Rmat \Xvect\pri_j) (\Yvect\pri_j - \rho \Rmat \Xvect\pri_j)\tp,
\end{align}
with the following subsitutions:
\begin{align}
\label{eq:X-centers}
 \overline{\Xvect} &= \frac{ \sum_{j=1}^{J} \overline{w}_j \Xvect_j}{\sum_{j=1}^N \overline{w}_j}, \quad
  \Xvect\pri_j =\Xvect_j - \overline{\Xvect}, \\
  \label{eq:Y-centers}
 \overline{\Yvect}  &= \frac{ \sum_{j=1}^{J} \overline{w}_j \Yvect_j}{\sum_{j=1}^N \overline{w}_j}, \quad
\Yvect\pri_j = \Yvect_j - \overline{\Yvect}.
\end{align}
By substituting \eqref{eq:opt-translation-robust} in \eqref{eq:expectation-student} and using the notations above, the rotation is estimated by minimization of:
\begin{equation}
\label{eq:opt-rotation-robust}
\Rmat^{\star} = \argmin_{\Rmat} \; \frac{1}{2}  \sum_{j=1}^{J} ( \overline{w}_j \| \Yvect\pri_{j}- \rho \Rmat \Xvect\pri_j \|^{2}_{\Sigmamat}).
\end{equation}
The minimizer \eqref{eq:opt-rotation-robust} yields a closed-form solution for an isotropic covariance, i.e. $\Sigmamat=\sigma\Imat$. In the general case, however, one has to make recourse to nonlinear minimization. It is practical to represent the rotation with a unit quaternion and to minimize \eqref{eq:opt-rotation-robust} using a sequential quadratic programming method \cite{bonnans2006numerical}. 
The parameter $\mu$ is updated by solving the following equation, where $\Psi(\cdot)$ is the digamma function:
\begin{equation}
\label{eq:mu-digamma}
\mu = \Psi^{-1}\left( \Psi(a) -\frac{1}{n} \sum_{j=1}^{J} \log b_j  \right).
\end{equation}
This yields Algorithm~\ref{algo:em-student}:

\begin{algorithm}[h!]
 \caption{\label{algo:em-student} Robust estimation of the rigid transformation between two 3D point sets.}

 \KwData{Centered point coordinates $\Xvect_{1:J}\pri$ and $\Yvect_{1:J}\pri$}

 \textbf{Initialization of $\thetavect^{\mathrm{old}}$}: Use a closed-form solution for the rotation to evaluate $\Rmat^{\mathrm{old}}$ and $\rho^{\mathrm{old}}$;  evaluate $\Sigmamat^{\mathrm{old}}$ using \eqref{eq:covariance-student}. Provide $\mu^{\textrm{old}}$ \;
 
 \While{$\|\thetavect^{\mathrm{new}} - \thetavect^{\mathrm{old}}\|>\epsilon$}{
 \textbf{E-step}: evaluate $a^{\mathrm{new}}$ and $\bvect_{1:N}^{\mathrm{new}}$ using \eqref{eq:gamma-posterior-ab} with $\thetavect^{\mathrm{old}}$, then evaluate $\overline{\wvect}_{1:N}^{\mathrm{new}}$ using \eqref{eq:weight-expectation} \;
 
 Update the centered coordinates  \;

 \textbf{M-scale-step}:   Evaluate $\rho^{\mathrm{new}}$ using \eqref{eq:opt-scale-robust}\;
 
 \textbf{M-rotation-step}: Estimate $\Rmat^{\mathrm{new}}$  via minimization of \eqref{eq:opt-rotation-robust} \;
 
 \textbf{M-covariance-step}:  Evaluate $\Sigmamat^{\mathrm{new}}$ using \eqref{eq:covariance-student} \;
 
  \textbf{M-mu-step}:   Evaluate $\mu^{\mathrm{new}}$ using \eqref{eq:mu-digamma} \;
  
  $\thetavect^{\mathrm{old}} \leftarrow \thetavect^{\mathrm{new}}$\;
 }
 
  \textbf{Optimal translation}: Evaluate the translation vector using \eqref{eq:opt-translation-robust}\;
 
  \KwResult{Optimal scale $\rho^{\star}$, rotation $\Rmat^{\star}$, translation $\tvect^{\star}$, covariance $\Sigmamat^{\star}$, and precisions  $\wvect_{1:N}$.}

\end{algorithm}

\section{Deformable Shape Model}
\label{app:statistical-shape}
In this appendix we summarize the deformable shape model that we use. The model is based on a training set $\mathcal{S}= \{\Svect_1, \dots, \Svect_m, \dots, \Svect_M\}$ of $M$ 3D face scans. Each 3D face scan is described with a triangulated mesh composed of $N$ 3D points, or vertices, $\Vvect_{1:N} = \{ \Vvect_1, \dots, \Vvect_N\}$, where the coordinates of each vertex are $\Vvect_n = (V_{n1}, V_{n2}, V_{n3})\tp$. Hence, a shape in $\mathcal{S}$ is described by the vector $\Svect$:
\begin{equation*}
\Svect=(V_{11}, V_{12}, V_{13}, \dots, V_{N1}, V_{N2}, V_{N3})\tp\in \mathcal{S} \subset \mathbb{R}^{3N}.
\end{equation*}
It is assumed that all the shapes are represented in the same coordinate frame, that they share the same number of vertices and that there is a one-to-one correspondence between their vertices, namely $\{\Vvect_{nm} \leftrightarrow \Vvect_{nm'}\}_{n=1}^{N}$, for any two shapes in the training set, $\Svect_m$ and $\Svect_{m'}$ and for each vertex index $n$. Notice, however, that the problem of finding a rigid alignment between $M$ shapes based on point-to-point correspondences is not trivial because of the presence of non-rigid shape deformations and of large shape variabilities.  In what follows we will treat each shape/face as a point set.
The covariance matrix associated with these $M$ registered shapes, $\{\Svect_m\}_{m=1}^{M}$, is:
\begin{figure*}[t!]
\centering
 \includegraphics[width=0.95\linewidth]{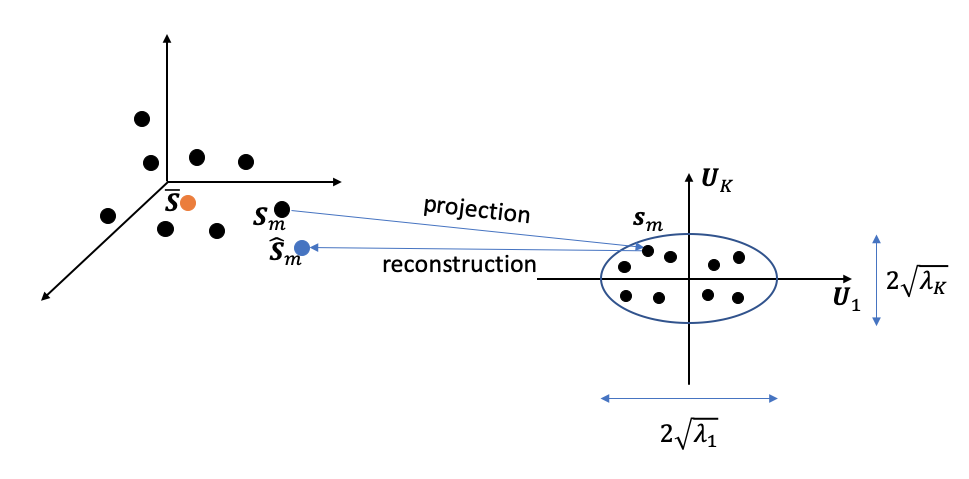}
\caption{\label{fig:active-shape} A training set of $M$ shapes (black dots), characterized by their mean (orange dot) and covariance, is projected onto the space spanned by the the principal eigenvectors of the shape covariance matrix, $\Uvect_1$ to $\Uvect_K$. The number $K$ of principal eigenvectors is selected such that the sum of the associated eigenvalues represents 95\% of the total variance. A shape $\Svect_m$ projects onto a shape embedding $\svect_m$. Conversely, shape $\hat{\Svect}_m$ (blue dot), reconstructed from $\svect_m$, approximates shape $\Svect_m$.}
\end{figure*}

\begin{align}
\label{eq:covariance-matrix}
\Sigmamat &= \frac{1}{M} \sum_{m=1}^M (\Svect_m - \overline{\Svect}) (\Svect_m - \overline{\Svect}) \tp,
\end{align}
where $\overline{\Svect} = ( \overline{\Vvect}_1\tp, \dots, \overline{\Vvect}_N\tp)\tp \in \mathbb{R}^{3N}$ is the mean shape with,
$\overline{\Vvect}_n = 1/M \sum_{m=1}^M \Vvect_{nm}$.
The covariance $\Sigmamat \in \mathbb{R}^{3N \times 3N}$ is a symmetric semi-definite positive matrix, hence its eigendecomposition writes
$\Sigmamat = \Umat \Lambdamat \Umat\tp$,
where $\Umat = \begin{pmatrix} \Uvect_1 & \dots & \Uvect_{3N}\end{pmatrix}\in\mathbb{R}^{3N \times 3N}$ is an orthogonal matrix and $\Lambdamat = \diag \begin{pmatrix} \lambda_1, \dots, \lambda_{3N}\end{pmatrix}$, with $\lambda_1 \geq \dots \geq  \lambda_{3N}\geq 0$. 
It is common practice to select the number $K$ of \textit{principal components}, or \textit{modes}, such that the sum of the $K$ largest eigenvalues represents 95\% of the total variance, $\sum_{k=1}^K\lambda_k = 0.95 \sum_{n=1}^{3N} \lambda_n$. We introduce the following \textit{truncated} matrices:
\begin{align*}
\widetilde{\Umat} &= \begin{pmatrix} \Uvect_1 & \dots & \Uvect_K\end{pmatrix} \in \mathbb{R}^{3N \times K}, \\
\widetilde{\Lambdamat} &= \diag \begin{pmatrix} \lambda_1, \dots, \lambda_{K} \end{pmatrix} \in \mathbb{R}^{K \times K},
\end{align*}
with the property $\widetilde{\Umat}\tp\widetilde{\Umat}=\Imat_{K}$.
Matrix $\widetilde{\Umat}\tp$ projects the centered shapes from $\mathbb{R}^{3N}$ onto $\mathbb{R}^{K}$:
\begin{equation}
\svect_m = \widetilde{\Umat}\tp (\Svect_m - \overline{\Svect}), %= \widetilde{\Umat} \tp \widetilde{\Svect}_m
\end{equation}
where $\svect_m\in\mathbb{R}^K$ is a low-dimensional embedding of $\Svect_m$.
Conversely, it is possible to reconstruct an approximation of $\Svect_m$, denoted  $\hat{\Svect}_m$,  from its low-dimensional embedding:
\begin{equation}
\label{eq:shape-reconstruction}
\hat{\Svect}_m = \overline{\Svect} + \widetilde{\Umat} \svect_m, %= \overline{\Svect} + \sum_{k=1}^{K} s_{mk} \uvect_k,
\end{equation}
which is a concatenation of:
\begin{equation}
\hat{\Vvect}_{nm} = \overline{\Vvect}_n + \Wmat_n \svect_m,
\end{equation}
where the $N$ block-matrices $\Wmat_n\in\mathbb{R}^{3\times K}$ are such that:
\begin{equation}
\label{eq:shape-modes}
\widetilde{\Umat} = \begin{pmatrix}
\Wmat_1 \\
\vdots \\
\Wmat_n \\
\vdots \\
\Wmat_N
\end{pmatrix}
\end{equation}

%Notice that this reconstruction is not exact because  we retained the largest $K$ eigenvalues. The error, between the ground-truth shape and the reconstructed one, is given by:
%\begin{equation}
%\| \Svect_m - \hat{\Svect}_m \| =  \| \sum_{k=K+1}^{3N} \svect_{mk} \uvect_k \|
%\end{equation}
The concept of \textit{statistical shape model} is summarized in Fig.~\ref{fig:active-shape}. A training set of $M$ shapes is embedded into a low-dimensional space spanned by the $K$ principal eigenvectors of the covariance matrix \eqref{eq:covariance-matrix} associated with the training set and is characterized by the $K$ principal (largest) eigenvalues, where an eigenvalue $\lambda_k$ corresponds to the variance along direction $\uvect_k$. Importantly, a \textit{valid} shape $\hat{\Svect}$
is a shape reconstructed from an
an embedding $\svect$ that lies inside an ellipsoid whose half axes  are equal to $\sqrt{\lambda_k}$, or:
\begin{equation}
\label{eq:shape-confidence}
\hat{\Svect} = \overline{\Svect} + \widetilde{\Umat} \svect, \; \textrm{s.t.} \; \svect\tp\widetilde{\Lambdamat}\inverse\svect \leq 1.
\end{equation}
This guarantees that the reconstructed shape belongs to the family of training shapes with 99\% confidence.
%$\| \svect \|^2_{\Lambdamat} \leq 1$ ($\| \svect \|^2_{\Lambdamat}= \svect\tp\Lambdamat\inverse\svect$ is the Mahalanobis distance).

\section{Robust Estimation of the Alignment Between a Deformable Shape and a 3D Point Set}
\label{app:robust-shape-fit}
%robust-shape-fit

In this appendix we extend the robust alignment method of Appendix~\ref{app:robust-alignment} to deformable shapes. The expression \eqref{eq:expectation-student} can be re-written as:

\begin{align}
\label{eq:expectation-student-reg}
\mathcal{Q}(\thetavect) &=  \frac{1}{2} \sum_{j=1}^{J} \big( \overline{w}_j \| \Yvect_j - \sigma \Qmat ( \overline{\Vvect}_j + \Wmat_j \svect) - \dvect \|^2_{\Sigmamat} + \log |\Sigmamat| \big) \nonumber \\
&+ \frac{\eta}{2} \svect\tp\widetilde{\Lambdamat}\inverse\svect,
\end{align}
where the statistical shape model is scaled, rotated, translated and deformed in order to be aligned with a set of landmarks, and where the
last term constrains the embedding $\svect$ to lie inside a 99\% confidence ellipsoid, i.e. \eqref{eq:shape-confidence}.

By taking the derivatives of \eqref{eq:expectation-student-reg} with respect to the parameters, i.e. $\thetavect = \{\sigma, \Qmat, \dvect, \Sigmamat, \svect,\mu\}$, and equating to zero, we obtain the following analytical expressions for the optimal values of the following parameters:
\begin{align}
\label{eq:opt-translation-robust-reg}
\dvect^{\star} &= \overline{\Yvect} - \sigma \Qmat \overline{\hat{\Vvect}},\\
\label{eq:opt-scale-robust-reg}
\sigma^{\star} & = \left( \frac{\sum_{j=1}^{J} \overline{w}_j \Yvect_j^{\prime\top}  \Sigmamat\inverse \Yvect\pri_j}{\sum_{j=1}^{J} \overline{w}_j (\Qmat \hat{\Vvect^{\prime}}_j )\tp  \Sigmamat\inverse (\Qmat\hat{\Vvect^{\prime}}_j)} \right) ^{\frac{1}{2}}, \\
\label{eq:covariance-student-reg}
\Sigmamat^{\star} &= \frac{1}{N} \sum_{j=1}^{J} \overline{w}_j (\Yvect\pri_j - \sigma \Qmat \hat{\Vvect\pri}_j) (\Yvect\pri_j - \sigma \Qmat \hat{\Vvect\pri}_j)\tp,
\end{align}
where similar notations as in Appendix~\ref{app:robust-alignment} and  Appendix~\ref{app:statistical-shape} were used, i.e \eqref{eq:X-centers}, \eqref{eq:Y-centers} and \eqref{eq:shape-reconstruction}. An identical nonlinear optimization problem for the rotation is obtained as well:
\begin{equation}
\label{eq:opt-rotation-robust-reg}
\Qmat^{\star} = \argmin_{\Qmat} \; \frac{1}{2}  \sum_{j=1}^{J} ( \overline{w}_j \| \Yvect\pri_{j}- \rho \Qmat \hat{\Vvect\pri}_j \|^{2}_{\Sigmamat}).
\end{equation}
Additionally, there is a closed-form solution for the optimal embedding:
\begin{equation}
\label{eq:opt-shape-robust-reg}
\svect^{\star} = \left( \sum_{j=1}^J \overline{w}_j \Amat_j\tp \Sigmamat\inverse  \Amat_j + \eta \widetilde{\Lambdamat}\inverse\right )\inverse \left( \sum_{j=1}^J \overline{w}_j \Amat_j\tp \Sigmamat\inverse \bvect_j \right),
\end{equation}
where the following notations were used: $\Amat_j= \sigma \Qmat \Wmat_j$ and $\bvect_j = \Yvect_j - \sigma \Qmat \overline{\Vvect}_j - \dvect$. The associated EM algorithm is similar to Algorithm~\ref{algo:em-student}; the most notable difference is that the M-step alternates between the estimation of the rigid parameters and the shape parameters, namely Algorithm~\ref{algo:em-student-shape}.

\begin{algorithm}[h!]
 \caption{\label{algo:em-student-shape} Robust fitting of a deformable shape model to a 3D point set.}

 \KwData{Centered landmark coordinates $\Yvect_{1:J}\pri$, mean shape $\overline{\Vvect}_{1:J}$ and shape reconstruction matrices $\Wmat_{1:J}$ }

 \textbf{Initialization of $\thetavect^{\mathrm{old}}$}: Set $\svect^{\mathrm{old}}=\zerovect$, use the closed-form solution \cite{horn1987closed}  to evaluate $\Qmat^{\mathrm{old}}$ and $\sigma^{\mathrm{old}}$;  evaluate $\Sigmamat^{\mathrm{old}}$ using \eqref{eq:covariance-student-reg}. Provide $\mu^{\textrm{old}}$. Compute the centered vertex coordinates $\hat{\Vvect\pri}_{1:J}$ \;
 
 \While{$\|\thetavect^{\mathrm{new}} - \thetavect^{\mathrm{old}}\|>\epsilon$}{
 \textbf{E-step}: evaluate $a^{\mathrm{new}}$ and $\bvect_{1:N}^{\mathrm{new}}$ using \eqref{eq:gamma-posterior-ab} with $\thetavect^{\mathrm{old}}$, then evaluate $\overline{\wvect}_{1:N}^{\mathrm{new}}$ using \eqref{eq:weight-expectation} \;
 
 Update the centered coordinates  \;

 \textbf{M-scale-step}:   Evaluate $\sigma^{\mathrm{new}}$ using \eqref{eq:opt-scale-robust-reg}\;
 
 \textbf{M-rotation-step}: Estimate $\Qmat^{\mathrm{new}}$  via minimization of \eqref{eq:opt-rotation-robust-reg} \;
 
 \textbf{M-covariance-step}:  Evaluate $\Sigmamat^{\mathrm{new}}$ using \eqref{eq:covariance-student-reg} \;
 
 \textbf{M-shape-step}: Evaluate $\svect^{\mathrm{new}}$ using \eqref{eq:opt-shape-robust-reg} \;
 
  \textbf{M-mu-step}:   Evaluate $\mu^{\mathrm{new}}$ using \eqref{eq:mu-digamma} \;
  
  $\thetavect^{\mathrm{old}} \leftarrow \thetavect^{\mathrm{new}}$\;
 }
 
  \textbf{Optimal translation}: Evaluate the translation vector using \eqref{eq:opt-translation-robust-reg}\;
 
  \KwResult{Optimal scale $\sigma^{\star}$, rotation $\Qmat^{\star}$, translation $\dvect^{\star}$, covariance $\Sigmamat^{\star}$, and precisions  $\wvect_{1:N}$.}

\end{algorithm}

\bibliographystyle{IEEEtran}
%\bibliography{citations,non-rigid-shape}

\end{document}